\documentclass[journal]{IEEEtran}

\usepackage{graphicx}
\usepackage{float}
\usepackage{stfloats}
\usepackage{array}
\usepackage{amsmath} 
\usepackage[utf8]{inputenc}
\usepackage[table]{xcolor}

\usepackage{amsfonts}

\begin{document}

% Linebreaks \\ can be used within to get better formatting as desired.
% Do not put math or special symbols in the title.
\title{Automatic Clustering for Unsupervised Risk Diagnosis of Vehicle Driving for Smart Road}

\author{Xiupeng~Shi,~Yiik~Diew~Wong,~Chen~Chai,~\IEEEmembership{Member,~IEEE,}~Michael~Zhi-Feng~Li,\\Tianyi~Chen,~Zeng~Zeng,~\IEEEmembership{Senior~Member,~IEEE}%

\thanks{This study was jointly sponsored by the National Key Research and Development Program of China (2018YFB1600502), the Chinese National Science Foundation (61803283), the “Chen Guang” project supported by Shanghai Municipal Education Commission and Shanghai Education Development Foundation (18CG17), and the Shanghai Municipal Science and Technology Major Project (No. 2021SHZDZX0100).}%
\thanks{Xiupeng Shi is with Institute for Infocomm Research, Agency for Science Technology and Research (A*STAR), 138632, Singapore (e-mail: shix@i2r.a-star.edu.sg).}%
\thanks{Zeng Zeng is with the School of Microelectronics, Shanghai University, Shanghai 201800, China; Institute for Infocomm Research, Agency for Science Technology and Research (A*STAR), 138632, Singapore (e-mail: zeng\underline{~}zeng@hotmail.com).}%
\thanks{Yiik Diew Wong and Tianyi Chen are with the School of Civil and
Environmental Engineering, Nanyang Technological University, Singapore
639798 (e-mail: cydwong@ntu.edu.sg; TIANYI002@e.ntu.edu.sg).}%
\thanks{Chen Chai is with the College of Transportation Engineering, Tongji University, Shanghai 201804, China (e-mail: chaichen@tongji.edu.cn).}%
\thanks{Michael Zhi-Feng Li is with the Nanyang Business School, Nanyang
Technological University, Singapore 639798 (e-mail: zfli@ntu.edu.sg).}
\thanks{Corresponding authors: Chen Chai.}
}

% headers
% \markboth{IEEE TRANSACTIONS ON INTELLIGENT TRANSPORTATION SYSTEMS}%
% {Shi \MakeLowercase{\textit{et al.}}: Automatic clustering for unsupervised risk diagnosis of vehicle driving for smart road}

% make the title area
\maketitle

\begin{abstract}
Early risk diagnosis and driving anomaly detection from vehicle stream are of great benefits in a range of advanced solutions towards Smart Road and crash prevention, although there are intrinsic challenges, especially lack of ground truth, definition of multiple risk exposures. This study proposes a domain-specific automatic clustering (termed AutoCluster) to self-learn the optimal models for unsupervised risk assessment, which integrates key steps of clustering into an auto-optimisable pipeline, including feature and algorithm selection, hyperparameter auto-tuning. Firstly, based on surrogate conflict measures, a series of risk indicator features are constructed to represent temporal-spatial and kinematical risk exposures. Then, we develop an unsupervised feature selection method to identify the useful features by elimination-based model reliance importance (EMRI). Secondly, we propose balanced Silhouette Index (bSI) to evaluate the internal quality of imbalanced clustering. A loss function is designed that considers the clustering performance in terms of internal quality, inter-cluster variation, and model stability. Thirdly, based on Bayesian optimisation, the algorithm auto-selection and hyperparameter auto-tuning are self-learned to generate the best clustering results. Herein, NGSIM vehicle trajectory data is used for test-bedding. Findings show that AutoCluster is reliable and promising to diagnose multiple distinct risk levels inherent to generalised driving behaviour. We also delve into risk clustering, such as, algorithms heterogeneity, Silhouette analysis, hierarchical clustering flows, etc. Meanwhile, the AutoCluster is also a method for unsupervised data labelling and indicator threshold calibration. Furthermore, AutoCluster is useful to tackle the challenges in imbalanced clustering without ground truth or \textit{a priori} knowledge.
\end{abstract}

\begin{IEEEkeywords}
Automatic clustering, Unsupervised feature selection, Unsupervised data labelling, Anomaly detection, Risk indicator.
\end{IEEEkeywords}

\IEEEpeerreviewmaketitle

\section{Introduction}

\IEEEPARstart{E}{arly} risk diagnosis and effective anomaly detection play a key role in a range of advanced solutions towards Smart Road, especially with the development of autonomous and connected vehicles (CAV) and roadside sensing \cite{eskandarian2019research}. Smart road will add huge benefits and synergistic effects to standalone smart vehicles, which can enable safety capacity beyond ego vehicle sensing \cite{Mozaffari2020}. Roadside sensing and CAV provide the capability of data acquisition, however, data analysis for risk diagnosis and anomaly detection remain weak, due to the inherent challenges of lack of risk definition, or no ground truth \cite{Zhu2018, Nallaperuma2019}. There is a perennial quest to develop measures to early-identify anomaly and risk potentials from generalised traffic conditions, especially based on easy collectable data.

For the purposes of early diagnostics and targeted countermeasures from root-causes, risk detection aims to uncover a wider scope of risk hierarchy and anomaly scenarios inherent in generalised non-crash traffic flow \cite{shi2019feature}. However, there are two main intrinsic challenges: (1) consensus definition of risk levels is not straightforward to determine, and (2) ground truth about risk degrees are lacking and expensive to obtain \cite{Shi2020}. Accident records have long been used as the basis to assess risk levels, but they are generally collected after the accident events. There are more risk-related conditions in near-misses, when compared with crash occurrences and insurance claim records \cite{Shi2018}. Surrogate measures of vehicle conflicts are well accepted as an effective way for risk evaluation \cite{Chai2015}, but they are based mostly on simulation and designed experiments \cite{chai2014micro}. In naturalistic driving study (NDS), near-crash incidents are typically tagged based on certain kinetics scenarios, such as rapid evasive manoeuvres \cite{Perez}. There are many viewpoints to describe risk conditions, and a comprehensive measure or consensus is still lacking. Besides, reliable thresholds for multiple risk hierarchy are difficult to determine.

A way of risk detection is by finding patterns that do not directly conform to expected safety \cite{Siami2020}. Clustering has an advantage in discovering data patterns from multiple dimensions, which is promising to detect outliers as the risk and anomaly instances, under the premise that majority instances are safe and normal. Notwithstanding, given the unavailability of ground truth labels, reliable clustering evaluation is problematic, which leads to a series of challenges such as suitability of algorithm selection, hyperparameter setting like the number of clusters, feature selection. Besides, feature extraction for risk clustering is rarely explored. Furthermore, risk detection is a distinctly imbalanced problem, while noting that modelling and evaluation for clustering on imbalanced data are likely biased towards the majority, which may produce misleading results \cite{shi2019feature}. Solutions for imbalanced clustering have not been well investigated.

The main contributions of the work can be summarized as follows:

(1) We introduce an automatic clustering method to self-learn the best models of risk clustering based on given data, which is also a method of unsupervised data labelling.

(2) We propose a set of solutions to address the main challenges in imbalanced clustering without ground truth, such as AutoML considering multiple clustering notions, indicator-guided feature extraction and unsupervised selection.

(3) We diagnose the risk exposure potentials in generalised traffic conditions, and delineate a high-resolution risk profiling application for Smart Road, which extend the scope of risk analysis as relying on crashes.

The focus of this paper is to explore unsupervised risk assessment and driving anomaly detection from vehicle trajectory data. Section II reviews the literature on clustering and risk evaluation. Section III elaborates the methodology of automatic clustering. Section IV presents the analysis of risk clustering and data-driven insights. The final two sections cover discussion and conclusions.

\section{Literature Review}

\begin{table*}[!t]
\centering
\caption{Clustering notions and algorithms}
\label{tab:1}
\begin{tabular}{ p{10em} p{30em} p{20em} }
\hline
Type & Notion of clusters & Representative algorithms \\ 
\hline
Centroid-based & Represented by a central vector (e.g., centroid, mode); assign the objects to the nearest cluster centre by (dis)similarity functions; drawbacks such as pre-defined k values; cut-borders & k-Means++; Mini Batch k Means; k-modes \\ 

Connectivity-based & Defined as connected dense regions in the data space; similarity in the objects are more related to nearby objects than to objects far away; produce hierarchical clustering; not very robust towards outliers & BIRCH (balanced iterative reducing and clustering using hierarchies); Ward hierarchical clustering (Ward); average linkage clustering (ALC) \\ 

Density-based & Defined as areas of higher density than the remainder of the data set; objects in sparse areas treated as noise and border points to separate clusters; work well on arbitrary shapes & DBSCAN (density-based spatial clustering of applications with noise); OPTICS (ordering points to identify the clustering structure); Mean-shift \\ 

Distribution-based & Defined by objects belonging most likely to the same statistical distribution; can capture correlation and dependence between attributes; need to align with distribution assumptions & Gaussian mixture \\ 

Fuzzy-based & Based on the fuzzy likelihood of belonging, each object belongs to more than one cluster to a certain degree; produce soft (alternative) clustering & Fuzzy C Means (FCM) \\ 
\hline
\end{tabular}
\end{table*}

\subsection{Unsupervised Risk Assessment}

One main objective of risk diagnosis and anomaly detection is to accurately identify multiple risk hierarchy from within generalised non-crash traffic flow using data easy to collect such as vehicle trajectory, which implies the capabilities of reducing crashes by mitigating pre-crash risk conditions. Data-driven algorithms are vitally important as they make inferences based on comprehensive patterns as learned from real-world data; they are promising towards finding hidden and in-depth patterns beyond conventional techniques (e.g., expert rules based on surrogate safety indicators), and also facilitate the knowledge discovery in risk assessment study.

Risk clustering entails using algorithms to group the vehicles (driving behaviours) with similar risk patterns into the same clusters, and then estimates the risk level of each cluster by pattern decoding. This method can provide data-driven insights about risk exposures in the traffic flow, and acts as an unsupervised procedure to label risk levels of vehicles.

The key challenges in risk clustering pertain to: (1) select appropriate algorithms, (2) define useful features as the inputs, and (3) tune hyperparameters to deliver optimal performance. Different clustering algorithms have specific mechanisms to form clusters, which produce different cluster geometry and application cases. Feature design is the most important step for domain-specific clustering, which provides useful input for algorithm learning. The optimisation of hyperparameters usually needs to try out all possible values, and automation of end-to-end process can thus be of great value-add in reducing the chore inherent in the trial-and-error process. Besides, risk conditions are complex and dynamic, and an automated process is beneficial in configuring proper algorithms, features, and hyperparameters to deliver the best-quality clusters for an updated and specific dataset.

\subsection{Surrogate Risk Measures}

Accident events are generally unexpected and occur rarely, and traffic conflict techniques (TCT) are the most prominent and well-recognised methods to identify risk anomaly and hazards that may lead to a crash \cite{Zheng2014}. In TCT, a series of surrogate risk measures have been developed to distinguish between risk and safety by either the intensity of evasive actions or the proximity in time and (or) space. Reference \cite{Mahmud2017} provides a comprehensive review of 17 surrogate risk indicators. However, indicators are often designed under simplified assumptions, such as unchanged trajectory, constant speed and predefined deceleration rate. Besides, reasonable thresholds for multiple risk levels are not straightforward to define. Integrated use of various indicators is suggested to represent complex risk mechanisms \cite{Laureshyn}, given that specific measures provide different cues and underlying information. The method for comprehensive risk assessment is not easy, since each indicator has a specific viewpoint, and a consensual result by multi-dimension criteria is hard to achieve. Recently, reference \cite{Shi2018} have retrieved high-resolution pre-crash vehicle trajectory data from real-world accident cases, and evaluated the performance of various risk indicators.

\subsection{Clustering Methods}

The notion of a cluster, as found by different algorithms, varies significantly in its properties. Popular notions of clusters include groups with small distances between cluster members, dense areas in data space, intervals or particular statistical distributions, etc. Many clustering algorithms have been proposed to date \cite{Estivill-Castro2002, Rodriguez2014}, as described in TABLE I. Given there is no ground truth (i.e., information provided by direct observation) nor \textit{a priori} knowledge for validation, the algorithms for risk clustering are not straightforward to select, thus any promising algorithm can thereby deliver an explicable partition of the dataset. Besides, an algorithm that is designed for one kind of data patterns (e.g., geometric shape) may generally fail on a dataset that contains a radically different pattern \cite{Estivill-Castro2002}.

% TABLE 2　CVIs for clustering evaluation from internal quality

\begin{table*}[!t]
\centering
\caption{CVIs for clustering evaluation from internal quality}
\label{tab:2}
\begin{tabular}{ p{10em} p{15em} p{35em} }
\hline
Metrics & Estimation & Description \\ 
\hline
SI & $\frac{1}{N} \sum_{i=1}^{N} \frac{b(i)-a(i)}{\max \{a(i), b(i)\}}$ &	Range [-1, 1]; near +1 indicates that the sample is far away from the neighbouring clusters; 0 means the sample is on or very close to the decision boundary between two neighbouring clusters \\

DB & $\frac{1}{k} \sum_{i=1}^{k} \max _{i \neq j} \frac{\sigma_{i}+\sigma_{j}}{d\left(c_{i}, c_{j}\right)}$ &	The average similarity between each cluster $i$, and its most similar one $j$; $\sigma_{x}$ is cluster diameter, by average distance of all elements in cluster $x$ to centroid $c_x$; $d(c_{i}, c_{j})$ is distance between cluster centroids\\

CH & $\frac{\operatorname{tr}\left(B_{k}\right)}{\operatorname{tr}\left(W_{k}\right)} \times \frac{N-k}{k-1}$ & The ratio of the sum of between-cluster dispersion $B_k$ and of within-cluster dispersion $W_k$ for all clusters, where $B_k=\sum_{q=1}^{k} n_{q}(c_{q}-c_{E})(c_{q}-c_{E})^{T}$ and $W_k=\sum_{q=1}^{k} \sum_{x \in C_{q}}(x-c_{q})(x-c_{q})^{T}$\\
\hline
\end{tabular}
\end{table*}

\subsection{Class Imbalance}

The occurrence of risk conditions (i.e., the minority class) is usually much lower than the number of safety instances (i.e., the mass majority class). Incorrect assignment of risk instances into a safety class entails a great misclassification cost \cite{elkan2001foundations}. Machine learning on imbalanced data is challenging, since algorithms are generally driven by global optimisation, which is likely biased towards the majority, and the minority class might be wrongly ignored as noise or outliers \cite{Diez-Pastor2015}. Moreover, there are observed intrinsic challenges in class imbalance, such as: (1) presence of small disjuncts, (2) lack of density and information, (3) problem of overlapping between classes, (4) non-obvious borderline instances for the distinction between positive and negative classes, and (5) the identification of noisy data, among others \cite{Beyan}. In supervised learning, there are several tactics to reduce the impacts of class imbalance, such as data under-sampling and/or oversampling, cost-sensitive loss functions \cite{Lopez2013}, \cite{Diez-Pastor2015}. However, the techniques to handle unsupervised learning on imbalanced data is not well investigated. Yet risk clustering is to find convincing and useful partitions to retrieve the imbalanced classes.

\subsection{Clustering Evaluation}

The quality of clustering can be generally measured by two ways, namely, external validation and internal validation. External validation is by comparing the clustering result to the ground truth or well-defined reference. However, in most real applications, one can hardly claim that a complete knowledge of ground truth is available or valid. Thus, external evaluation is mostly used for synthetic data and tuning models.

Internal validation is more realistic in many real-world scenarios. Several clustering validity indices (CVIs) have been proposed to give quality estimates of intrinsic properties about the found clusters, and well-known CVI metrics include silhouette index (SI), Calinski-Harabasz index (CH) (also known as the Variance Ratio Criterion), and Davies–Bouldin index (DB) \cite{de2015recovering, Van}, as listed in TABLE II. According to general concepts about good clustering, the SI and CH scores are higher when the structure of clusters is dense and well separated, while a lower DB score relates to a better separation between clusters. However, a common disadvantage is that these CVIs estimate the global quality, which is highly prone to bias towards the majority clusters. Thus, a CVI that considers the detailed quality of each cluster is more reliable for imbalanced clustering.

\section{Methodology}

\subsection{Automatic Clustering}
Aiming at addressing the main challenges in imbalanced clustering without ground truth, an automatic clustering technique (termed AutoCluster) is developed to configure optimal models for unsupervised risk assessment and driving anomaly detection for Smart Road, as shown in Fig. 1. 

The AutoCluster integrates key steps of clustering into an auto-optimisable pipeline, including risk feature extraction and unsupervised selection, algorithm auto-selection, hyperparameter auto-tuning, among others, as depicted in Fig. 2. A portfolio of basic algorithms (as described in TABLE I) is shortlisted in the AutoCluster. The most promising clustering solutions are self-learned based on Bayesian optimisation, as described in Section III.B. The clustering inputs, 
including the construction of risk indicator features and unsupervised feature selection, are elaborated in Section III.C and III.D, respectively. Finally, the optimal clusters are evaluated, and the labels about distinct risk clusters are interpreted based on pattern decoding. Besides, several strategies are designed to reduce the impact of class imbalance for clustering.

\begin{figure}[!t]
\centering
\includegraphics[width=3.4in]{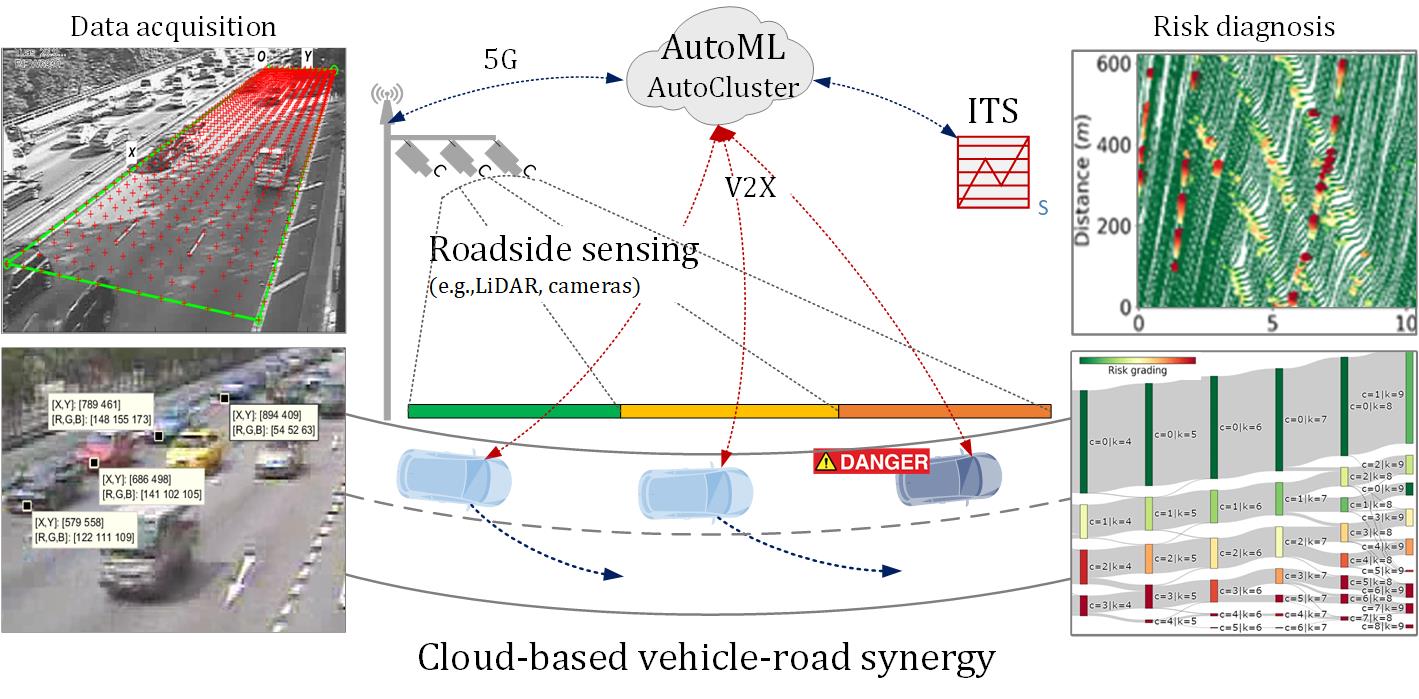}
\caption{AutoCluster for smart road.}
\label{fig1}
\end{figure}

\begin{figure}[!t]
\centering
\includegraphics[width=3.3in]{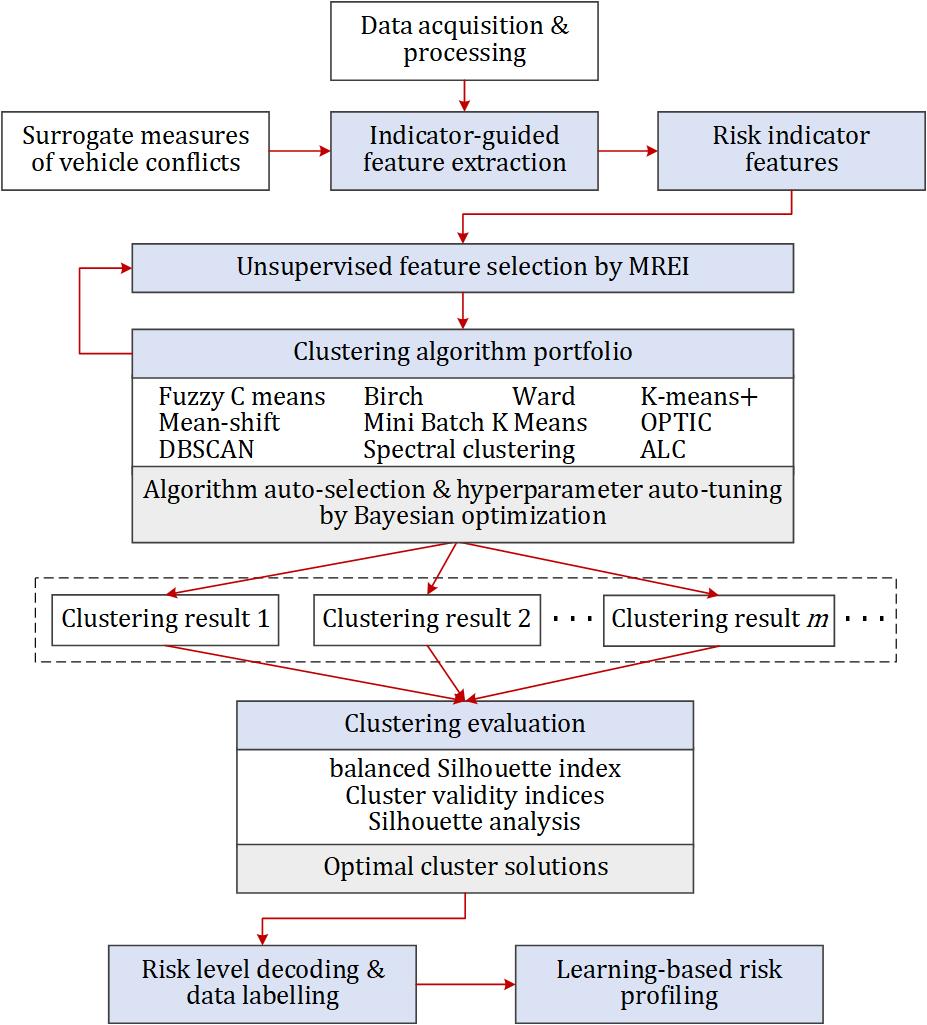}
\caption{Indicator-guided AutoCluster for unsupervised risk assessment.}
\label{fig2}
\end{figure}

\subsection{Bayesian-based Auto-Tuning of Model Selection and Hyperparameter Values}

\subsubsection{TPE-based Bayesian Optimisation}~

To find optimised clustering solutions, algorithm ($a$) selection and hyperparameter ($h$) tuning are conducted based on Bayesian optimisation, which is efficient for the tedious black-box tuning. A probabilistic surrogate model $p(y\mid x)$ of the objective function $f(x)$ is constructed to map input values $x\in M(a,h)$ to a probability of the loss, making it easier to optimise than the actual $f(x)$ \cite{shahriari2015taking}. By reasoning from past search results, next trials can concentrate on more promising ones, which can reduce total trials while finding a good optimum. 

The Bayesian-based auto-tuning of model selection and hyperparameter values is represented as:
\begin{equation}
    x^{*} =\arg \max _{x \in M(a, h)} f(x),
\end{equation}
\begin{equation}
\begin{split}
      EI_{y^{*}}(x) &=\int_{-\infty}^{y^{*}}\left(y^{*}-y\right) p(y|x) d y \\ 
      &\propto \left(\gamma+\frac{g(x)}{l(x)}(1-\gamma)\right)^{-1},
\end{split}
\end{equation}
\begin{equation}
    \gamma =p\left(y<y^{*}\right),
\end{equation}
\begin{equation}
    p(x|y) =\left\{\begin{array}{ll}
    l(x) & \text {if~} y<y^{*} \\
    g(x) & \text {if~} y \geq y^{*}
    \end{array}\right..
\end{equation}

Herein, $p(x|y)$ and quantile $\gamma$ are built based on Tree-structured Parzen Estimator (TPE) to produce a predictive posterior distribution of clustering models $M(a,h)$ over the performance of past results and form two non-parametric densities $l(x)$ and $g(x)$, which then guide the exploration of the model domain space \cite{Snoek, Bergstra}. The clustering models $x^{*}$ with the highest expected improvement $EI_{y^{*}}(x)$ are selected for next trials, which are expected to potentially minimise the loss function, namely, increase the performance. Hence, loss function design is key to efficient automatic clustering.

\subsubsection{Loss Function for AutoCluster}~

To retrieve reliable and versatile risk clusters, we design a loss function that considers the clustering performance in terms of internal quality $bSI$, inter-cluster variation $\sigma (S_k)$, and model stability $c_{V}$. The loss function is represented as:

\begin{equation}
    loss(x)=\left(S^{*}-bSI\right)+\lambda \cdot \sigma\left(S_{k}\right)+\varphi \cdot c_{V}, 
\end{equation}
\begin{equation}
    (a^{*}, h^{*})=\arg \min _{x \in M(a, h)} loss(x).
\end{equation}

\noindent where $\lambda$ and $\varphi$ are the weights to adjust the impact of each loss term into the loss function.

Generally, performance across different $k$ (i.e., number of clusters) is not straightforward comparable, and a smaller $k$ tends to show a lower loss, thus the auto-clustering process could bias towards placing most of the tries on smaller $k$ values. Hence, $loss(x\mid k)$ is further adjusted to estimate the extra optimisation conditioned on the same $k$, represented as:

\begin{equation}
    loss(x |k)=\frac{loss(x)}{loss(k)},
\end{equation}

\begin{equation}
    loss(k)=\frac{1}{|x|} \sum_{x} loss(x |k).
\end{equation}

\noindent where $loss(k)$ is the average loss for all results with the same $k$ till current iteration. The ratio form is useful to compare the extra improvement of performance across various $k$ due to auto-tuning, and to find the optimal algorithms and related hyperparameter values for each $k$.

\textit{(a) Balanced Silhouette Index}

Firstly, for a more nuanced estimation of imbalanced clustering, balanced Silhouette Index ($bSI$) is proposed to consider the weighted internal quality of all found clusters. As mentioned in Section II.E, the sample Silhouette Coefficient $s(i)$ is a measure of how similar a single instance is to its own cluster (i.e., cohesion) compared to neighbouring clusters (i.e., separation) \cite{Rousseeuw1987}. The cluster Silhouette score ($S_k$) is the mean of the $s(i)$ for all instance grouped into the cluster $k$. $bSI$ returns the weighted mean of $S_k$ averaged over all clusters, with weights $w_k$ to balance the quality of each cluster in the performance evaluation, represented as: 

\begin{equation}
    S(i)=\frac{b(i)-a(i)}{\max \{a(i), b(i)\}},
\end{equation}

\begin{equation}
    a(i)=\frac{1}{\left|C_{x}\right|-1} \sum_{j \in C_{x}, i \neq j} d(i, j),\left|C_{x}\right|>1,
\end{equation}

\begin{equation}
    b(i)=\min_{k \neq i} \frac{1}{\left|C_{y}\right|} \sum_{j^{\prime} \in C_{y}} d\left(i, j^{\prime}\right),
\end{equation}

\begin{equation}
    S_{k}=\frac{1}{\left|C_{k}\right|} \sum_{i \in C_{k}} S(i), 
\end{equation}

\begin{equation}
    bSI=\frac{\sum_{k=1}^{n} w_{k} \cdot S_{k}}{\sum_{k=1}^{n} w_{k}}.
\end{equation}

For $i \in C_x$ (i.e., data point $i$ assigned into the cluster $C_x$), $a(i)$ is the mean distance $d(i,j)$ between $i$ and all other data assigned in the same cluster ${j \in C_x}$. $C_y$ denotes neighbouring cluster next best for instance $i$. 

The $s(i)$ is bounded between [-1, 1]. A high $s(i)$ indicates that the instance $i$ is better matched to its own cluster than neighbouring clusters. Near $+1$ denotes highly dense clustering, when $a(i) \ll b(i)$. $s(i)<0$ generally indicates that $i$ has been assigned to a wrong cluster, as a different cluster is more similar. $s(i)$ around zero means that the instance is on the decision boundary of two neighbouring clusters, which also indicates overlapping.

In Eq.5, $(S^*-bSI)$ indicates the performance is optimised by reducing the difference between the optimal value $S^*$  (when all $S_k=1$) and $bSI$.

\textit{(b) Inter-cluster Variation}

Secondly, inter-cluster variation $\sigma (S_k)$ is added to penalise the imbalanced quality among the minority and majority clusters. $\sigma (S_k)$ is measured by the amount of variation or dispersion of $S_k$ from $bSI$, represented as: 
\begin{equation}
    \sigma (S_k)=\left(\frac{1}{|k|} \Sigma_{k \in[1, n]}(S_k-bSI)^{2}\right)^{1/2}.
\end{equation}

When $w_k$ is equivalent, $\sigma (S_k)$ is the standard deviation of $S_k$. Thus, reducing $\sigma (S_k)$ helps to achieve a balanced quality, which generally leads to an improvement on the minority.

With $bSI$ and $\sigma (S_k)$, the auto-tuning is greatly facilitated in finding algorithms and hyperparameters with lower loss, which indicates the clusters are dense and well separated, and minority clusters are configured appropriately as a standard concept of well-defined clusters.

\textit{(c) Model Stability}

In addition, model stability is also considered to satisfy the requirement of repeatability, which is important in optimising next tries. Generally, most clustering approaches have the properties of using random initialisation and approximate optimisation to search and assign clusters, which may produce different results in multiple same runs  \cite{Bousquet2002}.

Herein, the coefficient of variation ($c_V$) is applied to measure model stability based on the difference of results $y_{t}^{(a, h)}$ across running multiple replicates $t \in [1,T]$ with random initialisations and perturbing by small changes of hyperparameters. A model setting $(a,h)$ has good stability with respect to the expected results $\mathbb{E}_{t \in[1, T]}(y_{t}^{(a, h)})$,  if $\forall t \in\{1, \ldots, T\}$ the following holds:

\begin{equation}
    c_{V}(a, h)=\frac{\frac{1}{|t|}\| y_{t}^{(a, h)}-\mathbb{E}_{t \in[1, T]}(y_{t}^{(a, h)}) \|_{n}}{\beta+\mathbb{E}_{t \in[1, T]}(y_{t}^{(a, h)})}<\tau, 
\end{equation}

\begin{equation}
    y_{t}^{(a, h)}=\left\{s(i),~bSI,~\sigma (S_k), \dots \right\}|_{t}^{(a, h)} ,
\end{equation}

\begin{equation}
    c_{V}(a)=\frac{1}{|h|} \sum_{h} c_{V}(a, h).
\end{equation}

Thus, the model setting $(a,h)$ will be kept in the Bayesian optimisation domain space being tuned, when $c_V(a,h)<\tau$, where $\tau$ is a small value. Results $y_t^{(a,h)}$ of multiple replicates can be estimated by $s(i)$, $bSI$, or other metrics. $c_V(a,h)$ may be sensitive to small changes in a ratio scale, especially when $E_{t \in[1, T]}(y_t^{(a, h)}) \ll 1$, thus $\beta$ is added to prevent the ratio to change significantly by enlarging the denominator. $c_V(a,h)$ is measured based on the relative standard deviation when $n=2$. When the number of replicates varies across models, absolute difference in percent (herein $n=1$) is suggested to be superior. The algorithm stability $c_V (a)$ is then estimated by the mean value of $c_V (a,h)$.

\subsection{Indicator-guided Feature Extraction}

\begin{figure*}[t]
\centering
\includegraphics[width=50em]{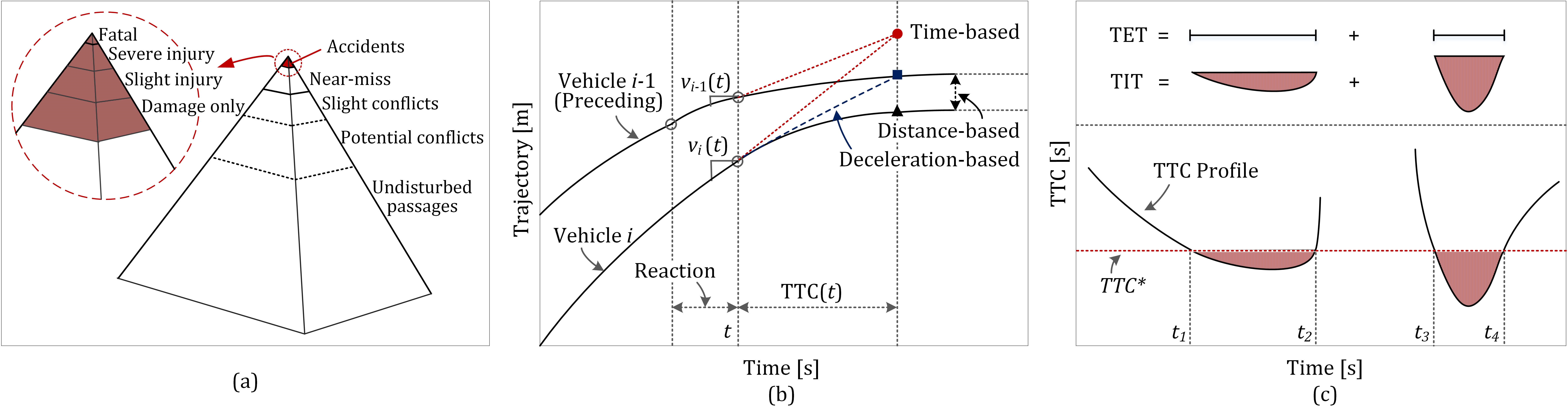}
\caption{Graphical representation of vehicle conflict indicators.}
\label{fig3}
\end{figure*}

Features play a key role in improving the quality of machine learning, which serves to bridge the gap between raw data and algorithm inputs \cite{Guyon2003}. For risk clustering, feature extraction is expected to derive and construct information that is effective to distinguish between risk and safety.  High interpretability is also critical for convincing and useful results. In addition, since risk detection is inherently an imbalanced problem, efficient features should be able to emphasise the learning on minority instances. Moreover, the features should be designed in line with convex clustering, to facilitate CVIs to deliver a fair and reliable estimation on clustering quality.

%Table 3　Risk features based on traffic conflict indicators

\begin{table*}[t]
\centering
\caption{Risk features based on traffic conflict indicators}
\label{tab:3}
\begin{tabular}{ p{5em} p{18em} p{7em} p{28em} }
\hline
Indicator &	Formulation	& Feature &	Explanation \\
\hline
TTC & $\frac{x_{i-1}(t)-x_{i}(t)-L_{i-1}}{v_{i}(t)-v_{i-1}(t)} ~~ \forall~v_{i}(t)>v_{i-1}(t)$ & $TTC.min$ & Minimum of average $TTC(t)$; $x_i(t)$ and $v_i(t)$ are the position and velocity of targeted vehicle $i$ at timestamp $t$, and $L_{i-1}$ is the length of preceding vehicle $i-1$\\ 

TET & $\sum_{t=0}^{N} \delta_{i}(t) \cdot \tau_{s c}$ & $TET.t1.max$; $TET.t2.max$; $TET.t3.max$ & Vehicle $TET(t)$ for a specific scope under TTC threshold $(TTC^{*})$; \{$t1:TTC^{*}=2$, $t2:TTC^{*}=3$, $t3:TTC^{*}=4$\}; $\tau_{sc}$ denotes small time intervals (e.g. $0.1$ second); ${\delta_{i}(t)}=1$, $\forall 0<TTC_{i}(t)<TTC^{*}$, else  ${\delta_{i}(t)}=0$\\ 

TIT & $\sum_{t=0}^{N}\left[T T C^{*}-T T C_{i}(t)\right] \cdot \tau_{s c}$ & $TIT.t1.max$; $TIT.t2.max$; $TIT.t3.max$ & Vehicle $TIT(t)$ for a specific scope under TTC threshold $(TTC^{*})$\\ 

DRAC & ${\frac{\left[v_{i}(t)-v_{i-1}(t)\right]^{2}}{x_{i-1}(t)-x_{i}(t)-L_{i-1}},~\forall~v_{i}(t)>v_{i-1}(t)}$ & $DRAC.max$ & Maximum value of vehicle $DRAC(t)$ for the defined scope\\ 

CPI & $\frac{1}{T} \sum_{t=0}^{N} P(DRAC_{i}(t)>M^{\alpha_{i}}) \cdot \tau_{s c}$ & $CPI.m1.max$; $CPI.m2.max$ & Vehicle $CPI(t)$ based on two kinds of MADR measurement ($m1$ and $m2$); MADR is specific for a given set of traffic and environmental attributes $\left\{\alpha_i\right\}$ \\ 

PSD & $\frac{R D}{v_{i}(t)^{2} / 2 d}$ & $PSD.min$; $PSD.mean$ & $d$ is acceptable maximum deceleration rate; $RD$ denotes remaining distance to the potential point of crash\\ 
\hline
\end{tabular}
\end{table*}

\subsubsection{Risk Indicator Features}~

Based on surrogate measures of vehicle conflicts, risk indicator features are extracted to represent driving risk exposures in terms of spatial-temporal and kinematical aspects. Previous studies have examined the feasibility of using risk indicators to assess pre-crash risk conditions \cite{Shi2018}. A hierarchy of risk is illustrated in Fig. 3(a), which can represent conflicts as early risk signals, avoidable risk conditions, near-miss, and crashes with various consequences.

Time to collision (TTC) is well-recognised and widely-used in conflict risk analysis \cite{Mahmud2017}. TTC is defined as time to a potential collision between two vehicles \cite{Horst1991}. Generally, risk conditions are flagged for any vehicle pair with a TTC value less than a given threshold. Based on TTC, Time Exposed TTC (TET) and Time Integrated TTC (TIT) have been further proposed to measure the severity by the duration and accumulation of time exposed in risk situations, respectively \cite{Minderhoud}. According to physical meanings, TTC-based indicators can reflect earlier risk signals.

Deceleration rate to avoid crash (DRAC) is recognised as a safety metric about performance in risk avoidance, which evaluates the braking requirement during a conflict \cite{Archer2005}. Based on DRAC, crash potential index (CPI) has been proposed to measure the likelihood of successful risk avoidance, which considers more factors such as maximum available deceleration rate (MADR) and time exposed to risk \cite{Cunto}. Hence, DRAC and CPI can describe avoidable risk conditions. 

Furthermore, when the maximum braking capability is executed, the proportion of stopping distance (PSD) measures the remaining distance to the potential point of crash. Figs. 3(b) and 3(c) depict the graphical representation of risk indicators. More information can be found in \cite{Shi2018}.

The feature extraction process entails the following steps: (1) select reliable surrogate risk indicators, and calculate the indicator values by a rolling window with a time interval, thus dynamic and accumulated risk can be measured; (2) sketch the key information of indicator time-series data by descriptive statistics (e.g., $min$, $max$, $mean$); and (3) for indicators involving predefined thresholds, several threshold values are set to cover various sensitivity. Herein, a total of 12 risk indicator features is constructed to represent various aspects of anomaly signals, as listed in TABLE \ref{tab:3}. The thresholds of TTC range from 2.0s to 4.0s, and two ways of CPI calculation (i.e., $m1$, and $m2$) are considered \cite{Shi2018}.

\subsubsection{Rectifier Function}~

A rectifier function $f(x)$ is designed to inhibit information of sufficient safety $x\in S$, by a weight $\phi \in [0,1)$, thus the learning attentions are pushed onto the risk domain, which can be regarded as a cost-sensitive strategy to improve imbalanced learning \cite{elkan2001foundations}. 
\begin{equation}
    f(x)=\left\{\begin{array}{cc}
\phi \cdot x, & \forall x \in S; \\
x, & \text { otherwise. }
\end{array}\right.
\end{equation}

Moderate rectifier thresholds are set to truncate the scales of sufficient safety, which is diluted to a limited range without under-estimation, so that the learning emphasises non-safety classes.

\subsection{Unsupervised Feature Selection}

Identifying the most useful features is also a necessary step. An optimal feature set is sufficiently informative for modelling, while reducing redundant, or useless information \cite{Zarshenas}. Benefits of key feature selection include simplification but pinpointing modelling, better interpretability \cite{Guyon2003}. However, classical feature selection methods are generally problematic in unsupervised learning, especially on imbalanced data.

This study develops an unsupervised feature selection method by elimination-based model reliance importance (EMRI). EMRI firstly evaluates the performance impact of an individual feature onto a certain clustering model based on feature elimination importance, and then aggregate the feature elimination importance to produce an model-agnostic importance.

\subsubsection{Feature Elimination Importance}~

The elimination importance $R_i^m$ of a feature $f_i$ for a clustering algorithm $m$ is measured based on the percentage change or difference ratio ($r_i^m$) in the performance before and after removing the feature, which is relative importance. The $R_i^m$ and $r_i^m$ are defined as: 

\begin{equation}
    r_{i}^{m}=\frac{S I_{i}^{m}-S I_{0}^{m}}{S I_{0}^{m}}=\frac{SI(c_{i}^{m}, \overbrace{F^{\backslash i}}^{n-1})}{SI(c_{0}^{m}, \underbrace{F_0}_{n})}-1, 
\end{equation}
\begin{equation}
    F^{\backslash i}=\left\{f_{1}, \ldots, f_{i-1}, f_{i+1}, \ldots, f_{n}\right\}, 
\end{equation}
\begin{equation}
    r_{i}^{m}=\mathbb{E}(r_{i}^{m}\mid_{n \rightarrow n-1})+R_{i}^{m},
\end{equation}
\begin{equation}
    R_{i}^{m} =r_{i}^{m}-\frac{1}{n} \sum_{i}^{n} r_{i}^{m} .
\end{equation}

Herein, a SI-based metric (i.e., $bSI$) is used to estimate the clustering performance. $SI_0^m$ denotes the performance calculated as conditioned on the initial feature set $F_0$ and algorithm $m$; $c_i^m$ denotes the cluster label generated by re-modelling with the remaining feature set $F^{\backslash i}$ after removing $f_i$.

The total change of performance after feature elimination is mainly due to two aspects, namely, the model-related performance change $\mathbb{E}(r_i^m \mid _{n \rightarrow n-1})$ and feature-related performance change $R_i^m$. Thereinto, $\mathbb{E}(r_i^m \mid _{n \rightarrow n-1})$ is linked to the inherent characteristics of the CVIs, which tend to have better scores with fewer features. Hence, after deducting the average change due to randomly removing any one feature, the extra value reflects the impact of the feature. Besides, the change ratio is used to be better comparable across different models, instead of absolute change. 

\subsubsection{Model-agnostic Feature Importance}~

The EMRI ($R_i$) is further designed as a model-agnostic feature importance to summarise $R_i^m$ across all well-performing models, which reflects on average the intrinsic value of a feature for the problem itself, in a comprehensive manner, rather than being conditioned on a prespecified model.

$R_i$ can be given based on mean value $R_i=\frac{1}{|m|} \sum_{|m|} R_i^m$ or the range of $R_i:\left[\min R_i^m, \max R_i^m\right]$.

The concerns are: (1) features are not with equal importance in different models which may lead to a Rashomon effect \cite{Fisher2019}; (2) features deemed as high importance in a bad model may be unimportant for a well-performing model; and (3) being hard to tune feature and model performance simultaneously or iteratively, since they are inter-conditioned.

The above method is straightforward wherein a feature is important (i.e., highly relevant and useful) if the performance (i.e., $bSI$) has a considerable decrease after the elimination, which implies the clustering highly relies on the feature information. A feature is less important if the performance is less changed or even improved after removing the feature, since an irrelevant feature only carries minimal impact while a redundant feature also has limited contribution due to the high correlation with other more important ones. Moreover, the quality may be diluted because additional noise or impurity is brought in from unimportant features. Besides, by re-modelling after the elimination, interactions among features are automatically considered.

\section{Analysis}
\subsection{Data}

AutoCluster is adaptable to a wide range of driving data that are collectable by CAV, roadside sensing, etc. As a test scenario, this study used the vehicle trajectory data from FHWA NGSIM (next generation simulation) programme. 

The data are of real-world generalised traffic flow during rush hours in the morning, collected from southbound US Highway 101, using several synchronised video cameras mounted on top of high buildings adjacent to the roadway \cite{halkias2006next}, \cite{Punzo}. The data of all vehicles in a 640-metre road segment for about 45 minutes are collected, such as vehicle trajectory, length, etc. The data acquisition resolution is 0.1 seconds, in relative units.

Data preprocessing uses the Savitzky-Golay filter to smooth out potential noise in data acquisition without much information loss due to cutting off peaks. After data cleaning, a total of 5,084 samples (vehicles) are used for modelling.

\subsection{Feature and Algorithm Selection}

AutoCluster divides a large scope of vehicles within a road segment into different groups, based on the pattern similarity as measured by risk indicator features. 

\subsubsection{Well-performing Algorithms}~

Firstly, to boost modelling efficiency and narrow the search space of auto-tuning, algorithms are preliminarily shortlisted by examining the model stability and average performance. The well-performing algorithms are then used in the subsequent feature selection and fine-tuning. 

A total of 10 clustering algorithms are tested, which consider various cluster notions and similarity measures (as discussed in Section II.C). As an important hyperparameter, the numbers of clusters ($k$), is predefined within a range of values from 3 to 9. Other hyperparameters are set using empirical values.

Herein, each algorithm is run 10 replicates with random initialisations to test the model stability by $c_V(a,k)$, in which the results $y_t^{(a,k)}$ are estimated by $bSI$ and $\sigma(S_k)$, the algorithm stability $c_V (a)$ is then estimated by the mean value of $c_V(a,k)$ averaged on $k$, as shown in TABLE \ref{tab:4}. The algorithm Mini Batch K Means is discarded, which has potentially better $bSI$ scores but vibrates a bit in model stability, since repeatability is an essential criterion of model selection.

The mean values of $bSI$, $\sigma(S_k)$ and other CVIs (e.g., $SI$, $CH$, $DB$; as stated in TABLE II) are plotted to compare the internal quality of various algorithms, as shown in Fig. 4. Down-arrow after the metric name indicates the lower value is better in performance. With the increase of $k$, the clustering quality trends to be reduced, especially when measured based on $SI$ and $CH$. Meanwhile, since there are more clusters thus more boundaries among adjacent clusters, the number of boundary samples and the likelihood of mis-cluster are also increased, as indicated by ${Sample~ Silhouette<0.05}$, in Fig. 4. The challenge is higher to have a superior retrieval resolution to gradually reveal more minority cases and differentiate higher-risk conditions.

Based on $bSI$ and $\sigma(S_k)$, 5 algorithms are shortlisted (as listed in TABLE V) as well-performing ones, which are ranked above the top $50^{th}$ percentile, as highlighted in non-grey colours in Fig. 4. Less-competitive ones are filtered out to reduce the search space of auto-tuning.

As shown in Fig. 4, difference CVIs have diverse suggestions on the best clustering. Basically, the CVIs measure the cluster compactness (e.g., instances in the same cluster should be similar) and separation (e.g., instances in different clusters should be dissimilar). Clusters with low intra-cluster distances (e.g., denser) and high inter-cluster distances (e.g., well separated) will have better CVIs scores. Obviously, the quality heterogeneity of individual clusters due to imbalanced data is not well presented by global CVIs such as $SI$ and $CH$. The $bSI$ and $\sigma(S_k)$ are rooted in the sample Silhouette coefficients, and adjusted to be more appropriate to explain a better assumption of imbalanced clustering.

% TABLE 4　Algorithm selection based on model stability

\begin{table}[!t]
\centering
\caption{Algorithm selection based on model stability}
\label{tab:4}
\begin{tabular}{p{16em} p{5em} p{5em}}
\hline
Algorithm &	$c_{V \mid bSI}$ & $c_{V \mid \sigma(S_k)}$ \\ 
\hline
1.Mini Batch K Means &	8.07\% &	16.79\% \\ 

2.k-means++ &	1.15\% &	3.72\% \\ 

3.Fuzz C Means &	0.26\% &	0.07\% \\ 

4.Spectral clustering & 0.04\% & 0.06\% 
\\ 

5.Mean Shift; 6.DBSCAN;~7.ALC; 
8.OPTICS; 9.Ward; 10.Birch & $<$0.01\% & $<$0.01\% \\ 
\hline
\end{tabular}
\end{table}

% Fig. 4. Algorithm selection based on performance of clustering internal quality 
\begin{figure}[!t]
\centering
\includegraphics[width=3.3in]{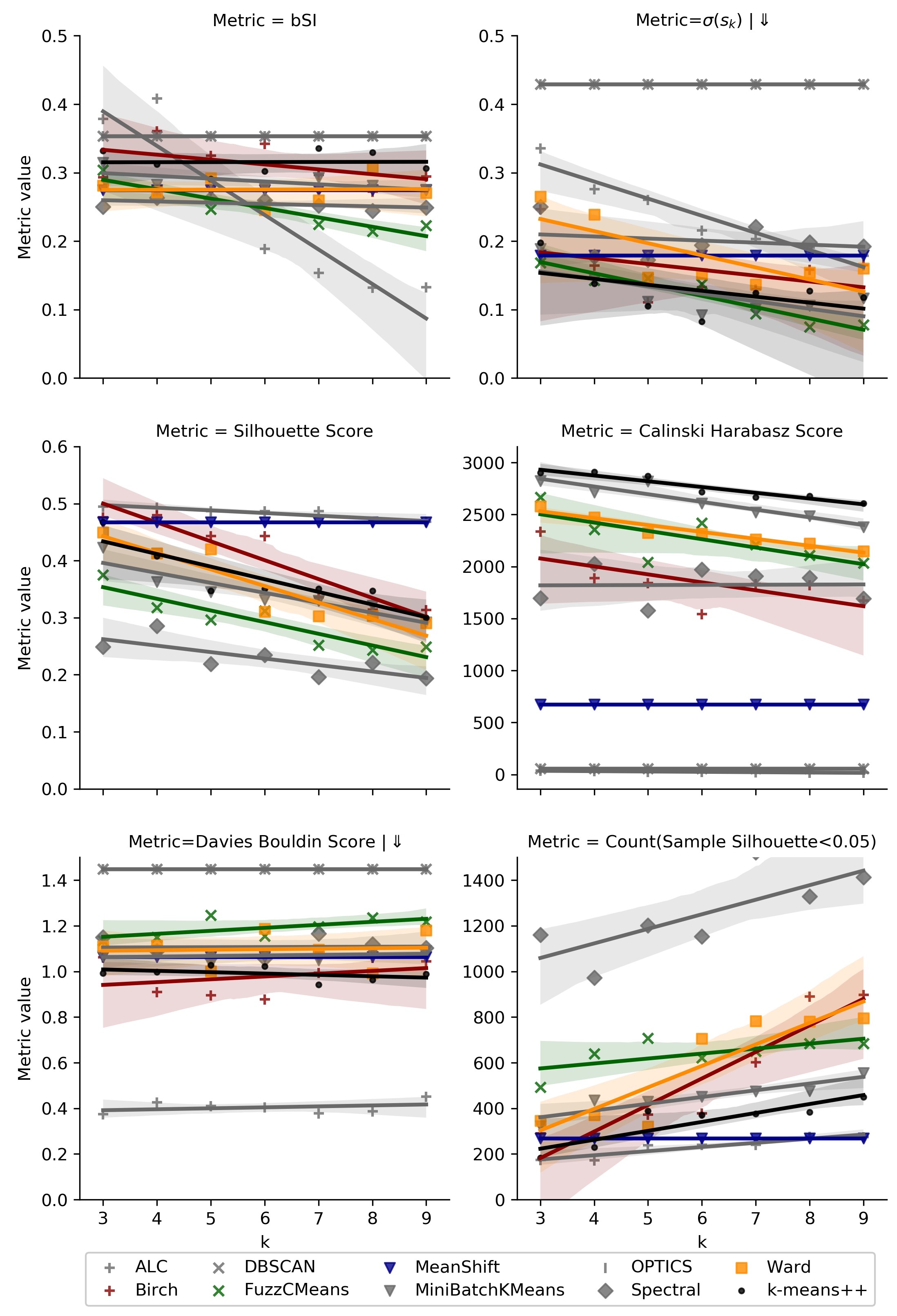}
\caption{Algorithm selection based on cluster internal quality.}
\label{fig4}
\end{figure}

\subsubsection{Feature Selection by EMRI}~

With the 5 well-performing clustering models, the feature importance is ranked based on EMRI. The EMRI ($R_i$) and overall performance change due to feature elimination (i.e., total change ratio $r_i^m$ after removing feature $f_i$) for each algorithm $m$ are listed in TABLE V. 

The $R_i$ measures the usefulness of a feature in the clustering problem. A smaller negative $R_i$ value indicates the feature of greater importance. Besides, $R_i>0$ indicates removing the feature will contribute to quality improvement. Thus, 8 features are selected from the initial set, which are $TET.t1.max$, $TIT.t1.max$, $TET.t3.max$, $CPI.m2.max$, $TIT.t3.max$, $TET.t2.max$, $TIT.t2.max$, $CPI.m1.max$. 

It is also found that algorithms have diverse sensitivity in randomly eliminating a feature, according to the model-related performance change ${\mathbb{E}(r_{i}^{m} \mid _{n \rightarrow n-1})}$ in TABLE V. Algorithms such as Ward and k-means++ are more tolerant of the changes in the inputs (e.g., feature elimination), that is being more robust to counter the disturbance and manipulation of data, as well as noise. Thus, subtracting ${\mathbb{E}(r_{i}^{m} \mid _{n \rightarrow n-1})}$ from overall performance change can offer a more reasonable estimation on the feature importance, which reveals the performance change caused by information loss of the feature itself. The model-agnostic EMRI summarises the feature elimination importance across all well-performing models, as the feature importance to the clustering problem, instead of being conditioned on a prespecified model. 

Besides, to improve clustering on imbalanced data, the rectifier function is applied to truncate the feature values within an effective and reasonable range, thus more learning attention is placed on the risk range. The rectifier thresholds are defined based on statistical distribution and physical meanings. Herein, the threshold is set as the 0.75 quantiles (i.e., sufficient safety) for the indicators whereby a higher value indicates safety. The up-bound of 7.95$s$ is set for $TTC$, and 2.0 for $PSD$. For physical meanings, 9.8$g$ ($m/s^2$) is set as the up-bound of the deceleration rate to truncate the out-of-range values of $DRAC$. Descriptive statistics of the extracted risk features are shown in TABLE V.

% TABLE 5　Unsupervised feature selection
\begin{table*}[t]
  \centering
  \caption{Unsupervised feature selection}
    \begin{tabular}{p{8em} p{25em} p{4em} p{4em} p{4em} p{4em} p{4em} p{5em}}
    \hline
    $f_i$ & \multicolumn{1}{l}{Feature description} & \multicolumn{5}{p{20em}}{$r_i^m$}  & \multicolumn{1}{c}{$R_i$} \\
          \cline{3-7} & \multicolumn{1}{l}{($mean$; $std$; $min$; $max$)} & \multicolumn{1}{l}{Birch} & \multicolumn{1}{l}{FCM} & \multicolumn{1}{l}{MS} & \multicolumn{1}{l}{Ward} & \multicolumn{1}{l}{k-means++} &  \\
    \hline
    \multicolumn{1}{l}{$TET.t1.max$} & \multicolumn{1}{l}{(0.104; 0.259; 0; 1)} & 0.318 & 0.396 & 0.546 & -0.174 & 0.006 & -0.668 \\
    \multicolumn{1}{l}{$TIT.t1.max$} & \multicolumn{1}{l}{(0.057; 0.189; 0; 1.723)} & 0.348 & 0.366 & 0.741 & -0.154 & 0.064 & -0.408 \\
    \multicolumn{1}{l}{$TET.t3.max$} & \multicolumn{1}{l}{(0.411; 0.444; 0; 1)} & 0.219 & 0.099 & 0.666 & -0.061 & 0.059 & -0.317 \\
    \multicolumn{1}{l}{$CPI.m2.max$} & \multicolumn{1}{l}{(0.008; 0.058; 0; 0.833)} & 0.203 & 0.363 & 0.733 & -0.091 & 0.078 & -0.241 \\
    \multicolumn{1}{l}{$TIT.t3.max$} & \multicolumn{1}{l}{(0.556; 0.806; 0; 3.723)} & 0.227 & 0.333 & 0.730  & -0.075 & 0.065 & -0.209 \\
    \multicolumn{1}{l}{$TET.t2.max$} & \multicolumn{1}{l}{(0.257; 0.387; 0; 1)} & 0.251 & 0.483 & 0.799 & -0.090 & 0.076 & -0.128 \\
    \multicolumn{1}{l}{$TIT.t2.max$} & \multicolumn{1}{l}{(0.230; 0.460; 0; 2.723)} & 0.237 & 0.334 & 0.750  & -0.052 & 0.056 & -0.126 \\
    \multicolumn{1}{l}{$CPI.m1.max$} & \multicolumn{1}{l}{(0.032; 0.141; 0; 1)} & 0.242 & 0.400   & 0.752 & -0.062 & 0.075 & -0.085 \\
    \multicolumn{1}{l}{\cellcolor[HTML]{DCDCDC} $DRAC.max$} & \multicolumn{1}{l}{(0.987; 1.033; 0.042; 9.8)} & 0.307 & 0.369 & 0.74  & -0.015 & 0.086 & \cellcolor[HTML]{DCDCDC} 0.143 \\
    \multicolumn{1}{l}{\cellcolor[HTML]{DCDCDC} $TTC.min$} & \multicolumn{1}{l}{(5.115; 2.263; 0.277; 7.95)} & 0.245 & 0.368 & 0.823 & -0.023 & 0.125 & \cellcolor[HTML]{DCDCDC} 0.177 \\
    \multicolumn{1}{l}{\cellcolor[HTML]{DCDCDC} $PSD.min$} & \multicolumn{1}{l}{(0.601; 0.319; 0.007; 2)} & 0.368 & 0.444 & 0.802 & 0.021 & 0.155 & \cellcolor[HTML]{DCDCDC} 0.532 \\
    \multicolumn{1}{l}{\cellcolor[HTML]{DCDCDC} $PSD.mean$} & \multicolumn{1}{l}{(1.217; 0.433; 0.195; 2)} & 0.543 & 0.546 & 0.673 & 0.145 & 0.241 & \cellcolor[HTML]{DCDCDC} 1.331 \\
    \hline
    \multicolumn{2}{c}{${\mathbb{E}(r_{i}^{m} \mid _{n \rightarrow n-1})}$} & 0.292 & 0.375 & 0.730  & -0.053 & 0.090  &  \\
    \hline
    \end{tabular}%
  \label{tab:5}%
\end{table*}%

\subsection{Bayesian-based Clustering Optimisation}

\subsubsection{Domain Space}~

Leveraging on Bayesian optimisation, AutoCluster screens out the best algorithms and hyperparameter settings from the domain space based on TPE. TPE chooses the next trials of algorithm and hyperparameter settings based on evaluating previous loss values, thus the auto-tuning can focus on more competitive ones. The domain space of algorithm selection and hyperparameter tuning are listed in TABLE \ref{tab:6}.

% TABLE 6　Algorithm and hyperparameter domain space

\begin{table*}[t]
  \centering
  \caption{Domain space for auto-tuning of algorithms and hyperparameters}
    \begin{tabular}{p{5em} p{12em} p{28em} p{10.5em} }
    \hline
    \multicolumn{1}{l}{Algorithm} & Hyperparameter & Description & Search domain \\
    \hline
    \multicolumn{1}{l}{Fuzz C Means} & Exponentiation of the fuzzy membership function ($m$) & Control the degree of fuzzy overlap between clusters, and fuzzy partition matrix & Uniform (1.1, 4.0) \\
    k-means++  & (1) $n$. initiations ($n$) & Number of initiations consecutive running with different centroid seeds to find the best output & Range (1, 100, 5) \\
          & (2) Type of core algorithm & Two variations of K-means algorithms to use & Choice (EM-style, Elkan) \\
    \multicolumn{1}{l}{Mean Shift} & Bandwidth quantile  & Quantile of the bandwidth used in the RBF kernel; 0.5 means using the median of all pairwise distances & Uniform (0.1, 1.0) \\
    \multicolumn{1}{l}{Ward} & $n$. neighbours for connectivity graph & Number of neighbours for each sample for the connectivity matrix; Ward minimises the variance of the clusters being merged & Range (5, 100, 5)  \\
    Birch & (1) Threshold  & The radius of the sub-cluster obtained by merging a new sample and the closest sub-cluster should be lesser than the threshold; lower values promote splitting & Uniform (0.1, 1.0)  \\
          & (2) Branching factor & Maximum number of CF (clustering feature) sub-clusters in each node; controls node splitting and sub-cluster re-distribution & Range (10, 100, 10)  \\
    \hline
    \end{tabular}%
  \label{tab:6}%
\end{table*}%

\subsubsection{Bayesian Auto-clustering}~

The $loss(x|k)$ and clustering models (i.e., both algorithms and hyperparameter $k$) versus the iterations are plotted to inspect the auto-tuning process, as shown in Fig. 5. The iteration of auto-tuning is set to be 1,000. 

From Fig. 5, the overall performance is improved (i.e., more tries of $loss(x|k)<1$ over time as expected, indicating that AutoCluster is trying to better the clustering results. $loss(x |k)$ represents the extra improvement than average performance during auto-tuning, which can facilitate Bayesian optimization for a unified comparison with various $k$ jointly, and avoid placing more tries on small $k$ (i.e., easier to cluster as indicated in Fig. 4).

After auto-tuning, the optimal clustering results for each $k$ are obtained, as well as corresponding algorithm and hyperparameter settings, as listed in Fig. 6 and TABLE \ref{tab:7}. Besides, the boundary instances are also counted, which are measured by $|s(i)|<0.05$.

It is found that: (1) there is no single algorithm that can deliver the best clustering results for all $k$ scenarios, and (2) each $k$ scenario has a specific optimal algorithm. An algorithm performs the best for a particular $k$ but maybe the worst for another. The algorithm selection and hyperparameter tuning are intercorrelated, and AutoCluster is efficient and promising to find the best solution from massive candidates.
 
%Figure 5　Bayesian-based clustering auto-tuning
\begin{figure}[!t]
\centering
\includegraphics[width=3.3in]{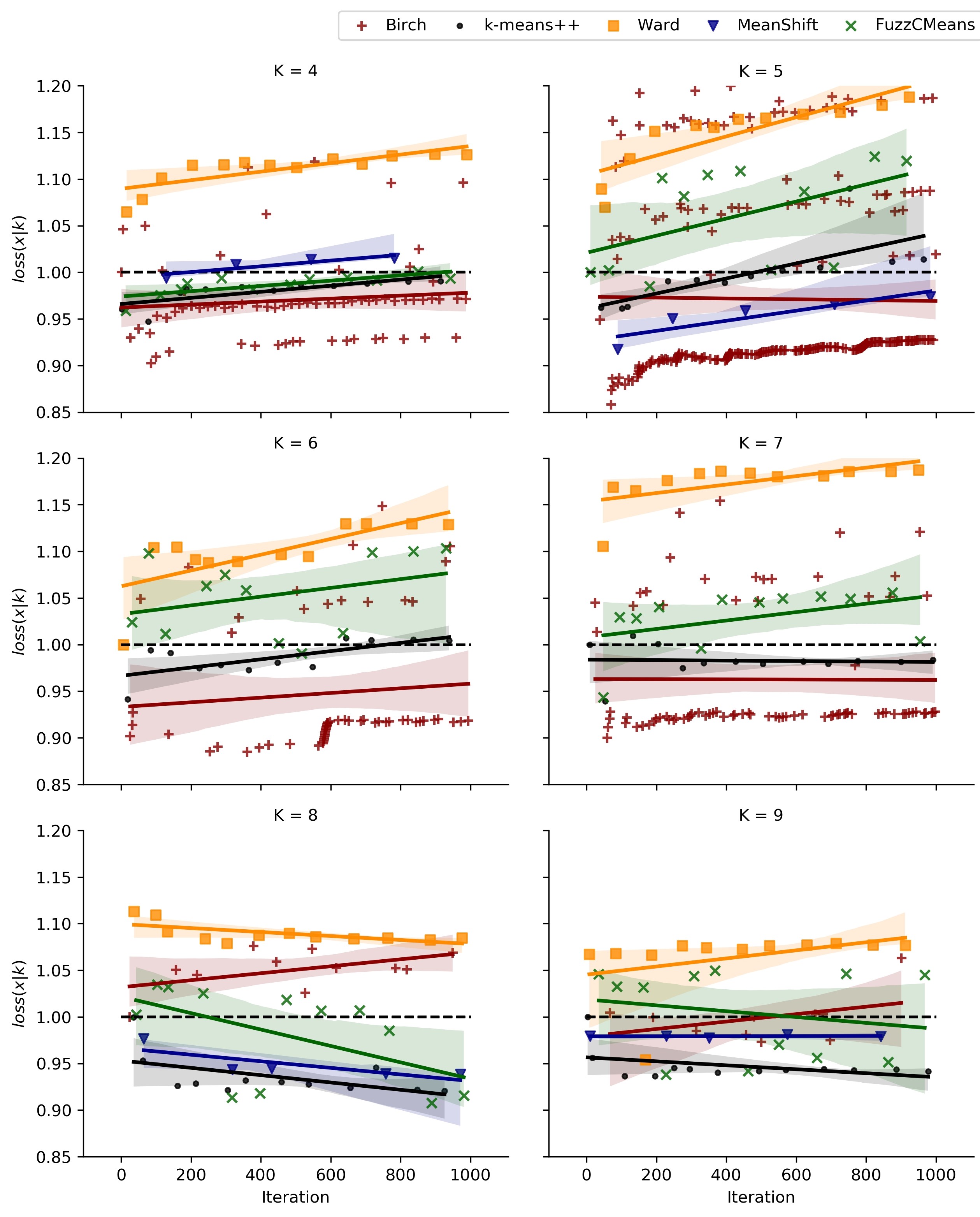}
\caption{Bayesian-based clustering auto-tuning.}
\label{fig5}
\end{figure}

%Table 7　Optimised clustering results and models

\begin{table*}[t]
  \centering
  \caption{Optimised clustering results and models}
    \begin{tabular}{rlrrrrrrrl}
    \hline
    $k$ & Optimal algorithm & $bSI$ & $\sigma(S_k)$ & Boundary instances & $loss(x |k)$ & $loss(x)$ & $loss(k)$ & Iteration & Optimal hyperparameters \\
    \hline
    3     & Birch & 0.591 & 0.181 & 74    & 0.908 & 0.499 & 0.550  & 433   & Branching factor: 50; Threshold: 0.2 \\
    4     & k-means++ & 0.554 & 0.235 & 71    & 0.986 & 0.564 & 0.572 & 608   & Type: Elka; n:1 \\
    5     & MeanShift & 0.541 & 0.213 & 75    & 0.917 & 0.565 & 0.616 & 89    & Quantile:0.4 \\
    6     & Birch & 0.535 & 0.202 & 76    & 0.908 & 0.566 & 0.623 & 585   & Branching factor: 50; Threshold: 0.4 \\
    7     & Birch & 0.501 & 0.200   & 85    & 0.928 & 0.598 & 0.645 & 998   & Branching factor: 80; Threshold: 0.4 \\
    8     & FuzzCMeans & 0.448 & 0.200   & 128   & 0.908 & 0.652 & 0.718 & 889   & m:1.8 \\
    9     & FuzzCMeans & 0.441 & 0.204 & 96    & 0.944 & 0.661 & 0.701 & 887   & m:1.2 \\
    \hline
    \end{tabular}%
  \label{tab:7}%
\end{table*}%

% Fig. 6　 Performance comparison of the best settings of each algorithm
\begin{figure}[!t]
\centering
\includegraphics[width=3.2in]{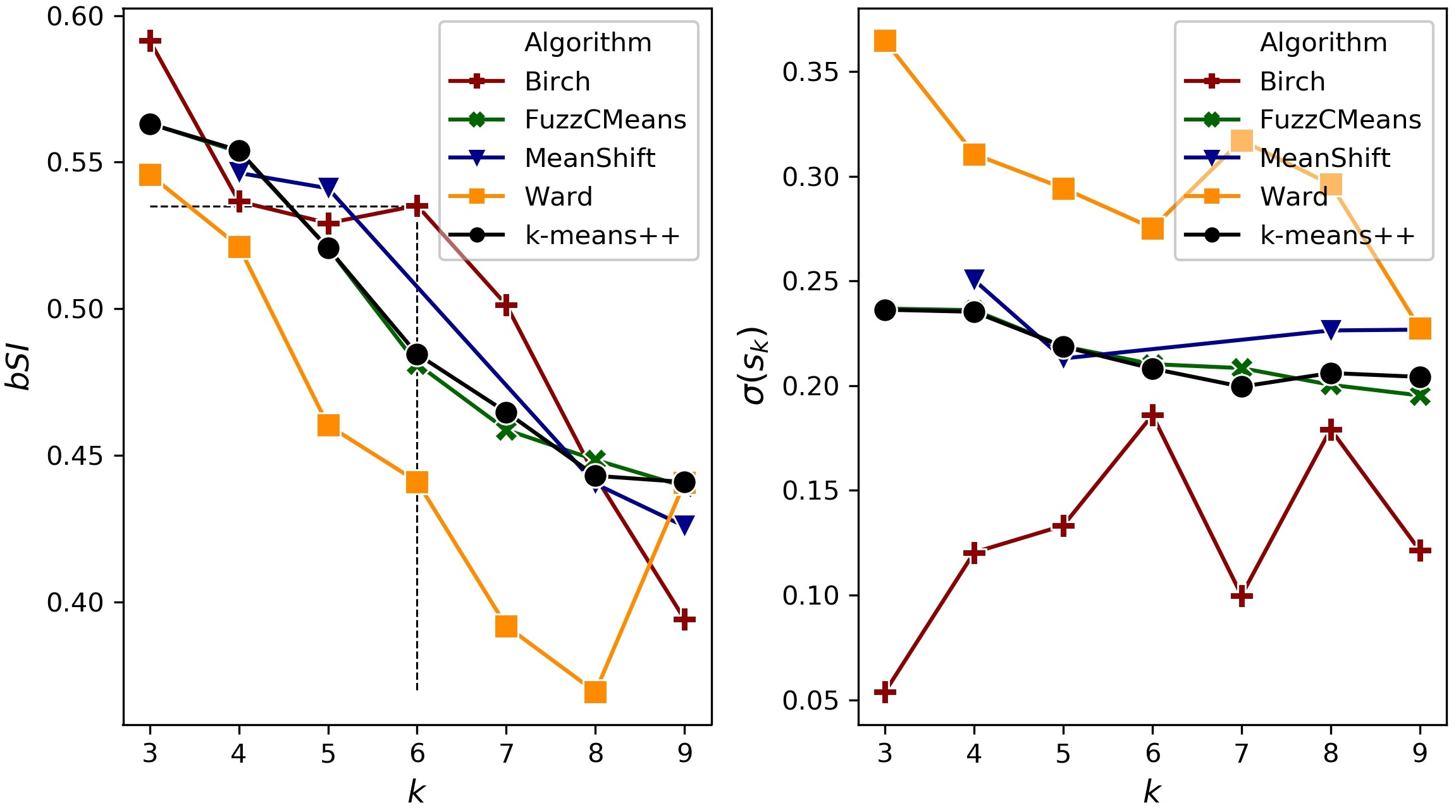}
\caption{Optimal performance for each algorithm.}
\label{fig6}
\end{figure}

\subsubsection{Algorithm Heterogeneity Analysis}~

Different clustering algorithms present diverse characteristics in the hyperparameter auto-tuning process. The heterogeneity is summarised based on $bSI$ by letter-value plots in Fig. 7. Letter-value plots display more detailed information about the tails behaviour and afford more precise estimates of corresponding quantiles, thus are more appropriate for imbalanced data.

From Fig. 7, it is found that: (1) k-means++ and Ward tend to have stable results with different hyperparameter values, which implies they are more straightforward to use without sophisticated tuning, (2) hyperparameter tuning has a greater impact on the performance of Birch and Fuzzy C means, and (3) the benefit of sophisticated hyperparameter tuning is obvious, especially for Birch. AutoCluster enables an integrated framework to automatically generate the best clustering results.

Besides, clustering ensemble presents a better overall performance in favour over the best stand-alone model, in both reliability and robustness. The end result can be constructed by a consensus function of majority voting, which iteratively adds the cluster labels generated by top-n algorithms, and assigns the majority as the end results. The decision boundary of the clustering ensemble is more convincing, even in the presence of noise and outliers, on average.
 
% Fig. 7　Algorithm heterogeneity analysis
\begin{figure}[!t]
\centering
\includegraphics[width=3.1in]{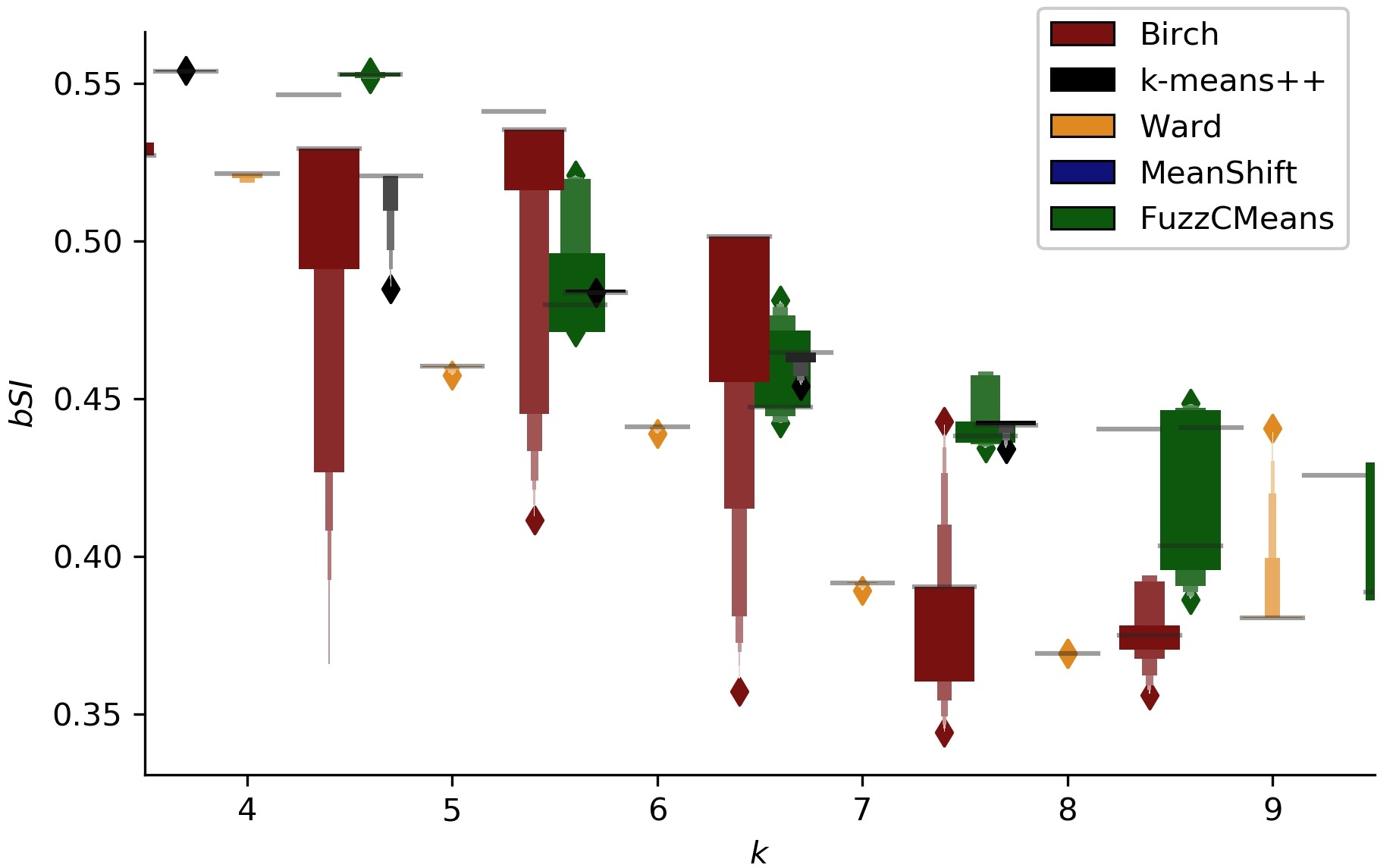}
\caption{Algorithm heterogeneity analysis.}
\label{fig7}
\end{figure}

\subsection{Risk Decoding and Data Labelling}
\subsubsection{Optimal Number of Clusters}~

The AutoCluster also contributes in determining an optimal number of clusters ($k$) without $a~priori$ knowledge. Hierarchical risk partitions formed by incremental $k$ are illustrated in Fig. 8. The Sankey diagram emphasises the flows of instances across corresponding clusters with different risk levels. The widths of the bands are linearly proportional to the total number of samples in each cluster.

Since the results across different $k$ are not directly comparable, Fig. 6 illustrates the trade-offs between cluster quality and retrieval resolutions, and $k=6$ is suggested to be the best result based on the elbow point method. The elbow point method is a widely used criterion, which entails plotting the cluster quality as a function of $k$, and picking the elbow of the curve as the optimal $k$ to fit. The elbow $k$ reflects a cutoff point of fitting and/or a signal of over-fitting, where diminishing returns are no longer worth additional costs, e.g., subdividing induces over-fitting.

With the increase of $k$, the retrieval resolution is increased, which facilitates to separate more minority cases from the majority, thus the partitions of highest risk levels are discovered. Meanwhile, a larger $k$ will encounter more instances that are placed in boundaries (as listed in TABLE \ref{tab:7}) or mis-clustered (as indicated by $Sample~ Silhouette<0.05$, in Fig.~4). 

\subsubsection{Silhouette Analysis}~

Based on sample $s(i)$, Silhouette analysis is conducted to further investigate the detailed separation distance and density between the formed clusters, as illustrated in Fig. 9. The total number in each cluster can be visualised from the thickness of the silhouette plot. Positive $s(i)$ values indicate that the samples have been assigned to the right cluster. 

The Silhouette plot provides a succinct graphical representation of how well each data point has been clustered. With reference to $bSI$ and $\sigma(S_k)$, the performance of minority clusters is consistent with the majority, without obvious quality differences as commonly presented in imbalanced learning. The Silhouette plot further supports the results of AutoCluster being reliable and of high-quality.
 
% Fig. 8　Hierarchical risk partitions and clustering flows by AutoCluster
\begin{figure}[!t]
\centering
\includegraphics[width=3.4in]{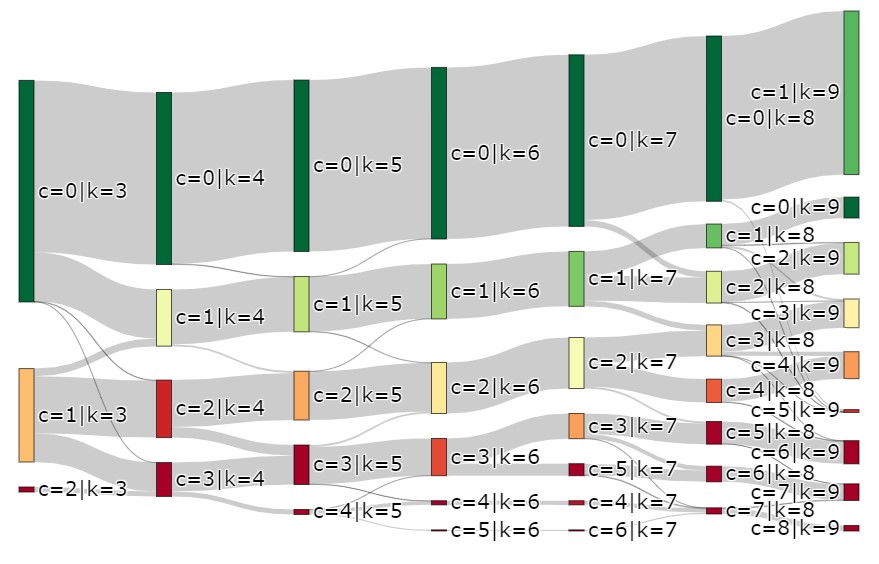}
\caption{Hierarchical risk partitions and clustering flows by AutoCluster.}
\label{fig8}
\end{figure}
 
% Fig. 9　Silhouette analysis for clustering evaluation from the sample data level
\begin{figure}[!t]
\centering
\includegraphics[width=3.3in]{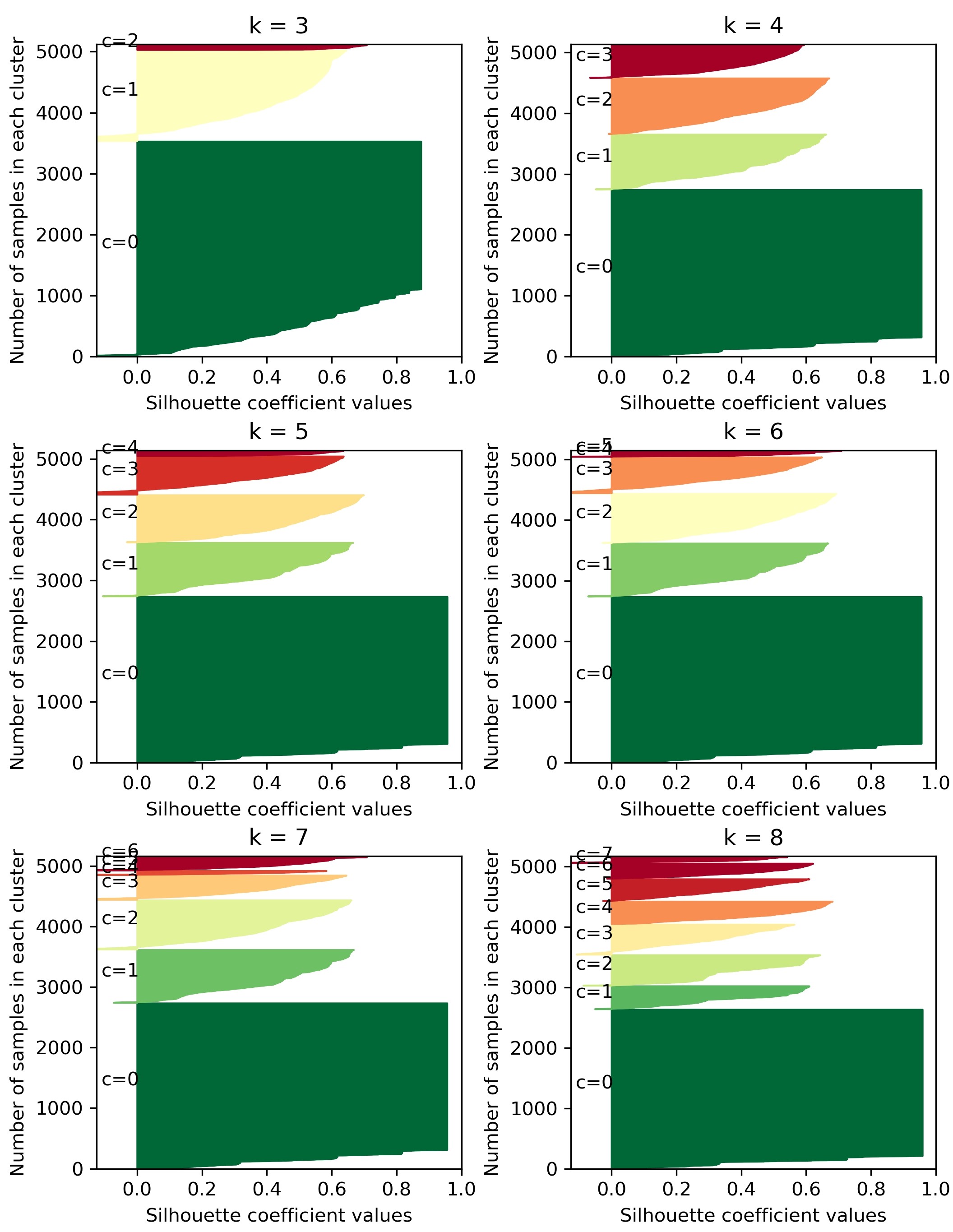}
\caption{Silhouette analysis for clustering evaluation from instance level.}
\label{fig9}
\end{figure}

\subsubsection{Risk Pattern Decoding}~

Vehicles with similar feature patterns are grouped into the same cluster, and each cluster has a respective distinct pattern. The selected key indicator features can facilitate to recognise and interpret the risk patterns of the formed clusters. The feature value distributions are summarised in Fig. 10 by letter-value plots. 

Based on the physical meanings and value distributions of the risk indicator features, the risk patterns are decoded with graded levels from safety to the highest severity, i.e., from 0 to 5 in Fig. 10.
 
% Fig. 10　Indicator feature value distribution for risk level decoding
\begin{figure}[!t]
\centering
\includegraphics[width=3.3in]{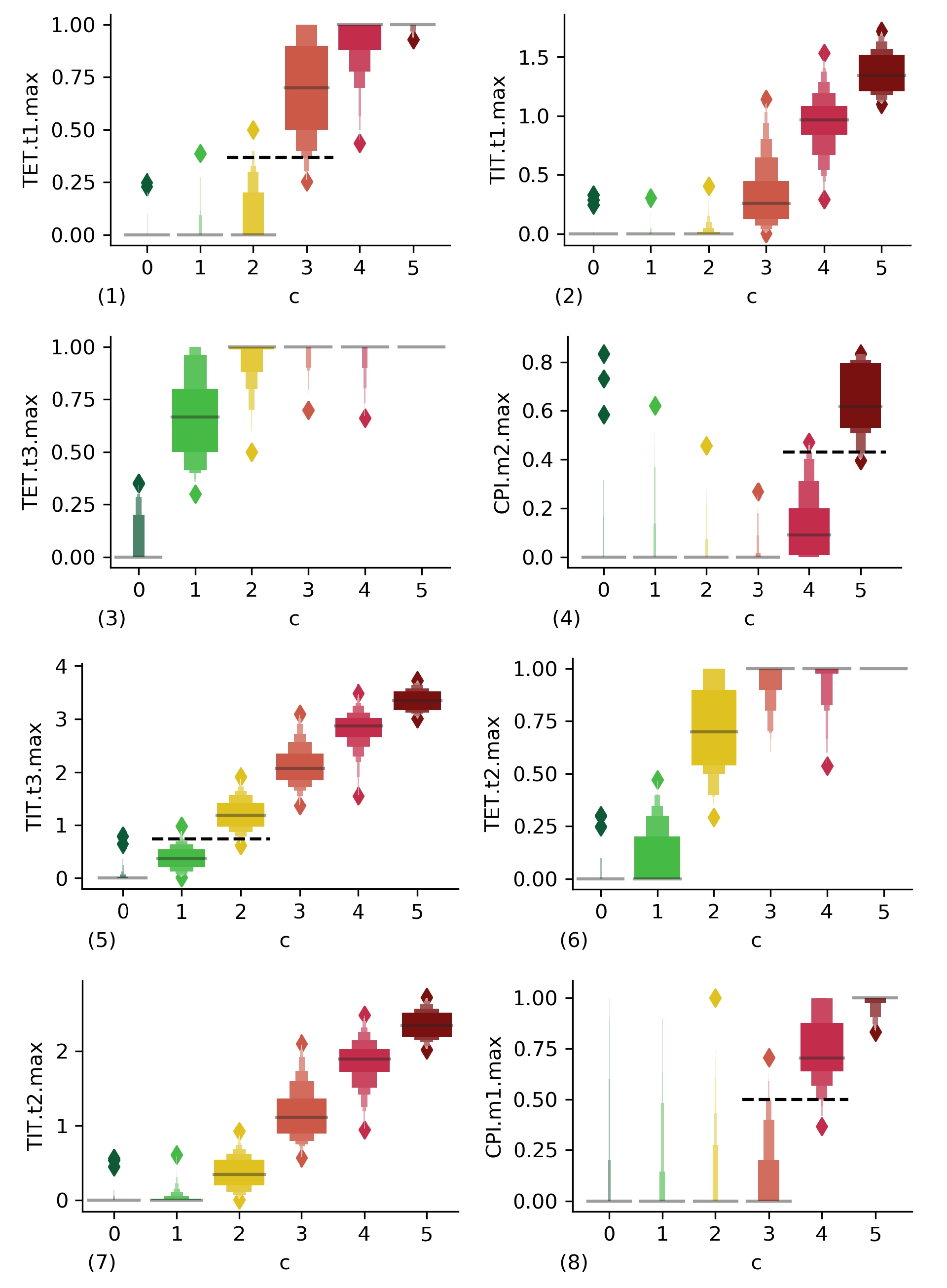}
\caption{Indicator feature value distribution for risk level decoding.}
\label{fig10}
\end{figure}

\subsubsection{Unsupervised Data Labelling}~

The clustering captures the position (i.e., risk level) of a data instance in the entire dataset based on multidimensional views of feature similarity, and a label about the belonged position can be assigned to the instance. 

Herein, according to the above risk pattern decoding, 6 risk levels can be defined. One annotation can be: safe level ($c=0$; with $2,725$ instances, about $53.6\%$ of total data; $indicators=0$); 3 risk levels from low to moderate ($c=1, ~2, ~3$; with instances of $872, ~810, ~591$, respectively; based on $TET~or~TIT>0$ with different thresholds sensitivity); and 2 high-risk levels ($c=4, ~5$; with $66$ and $18$ instances, respectively; based on $CPI>0$). The safe level indicates the lowest likelihood to be involved in vehicle crashes, and vice versa. Besides, imbalance ratio ($IR$) is commonly used to measure the class imbalance, a dataset is referred to as highly imbalanced if $IR\geq9$. Herein, the $IR$ is 32 in our dataset, calculated by $2,725/(66+18)$.

% TABLE 8　Hybrid indicators and thresholds for risk estimation

\begin{table*}[t]
  \centering
  \caption{Hybrid indicators and thresholds for risk estimation}
    \begin{tabular}{p{7em} p{7em} p{7em} p{7em} p{14em} p{7em}}
    \hline
          & Risk levels & $Mean$  & $Std$   & ($Min$, $Max$) & Threshold value \\
    \hline
    $TIT.t3.max$ & $c=1 \mid c=2$ & 0.38 $\mid$ 1.20 & 0.21 $\mid$ 0.28 & (0.05, 0.74) $\mid$ (0.73, 1.74) & 0.74 \\
    $TET.t1.max$ & $c=2 \mid c=3$ & 0.05 $\mid$ 0.69 & 0.11 $\mid$ 0.22 & (0, 0.37) $\mid$ (0.30, 1.0) & 0.37 \\
    $CPI.m1.max$ & $c=3 \mid c=4$ & 0.06 $\mid$ 0.74 & 0.14 $\mid$ 0.16 & (0, 0.50) $\mid$ (0.49, 1.0) & 0.50 \\
    $CPI.m2.max$ & $c=4 \mid c=5$ & 0.13 $\mid$ 0.65 & 0.14 $\mid$ 0.15 & (0, 0.44) $\mid$ (0.41, 0.83) & 0.44 \\
    \hline
    \end{tabular}%
  \label{tab:8}%
\end{table*}%

% Fig. 11　High-resolution dynamic risk detection and profiling

\begin{figure*}[!t]
\centering
\includegraphics[width=6in]{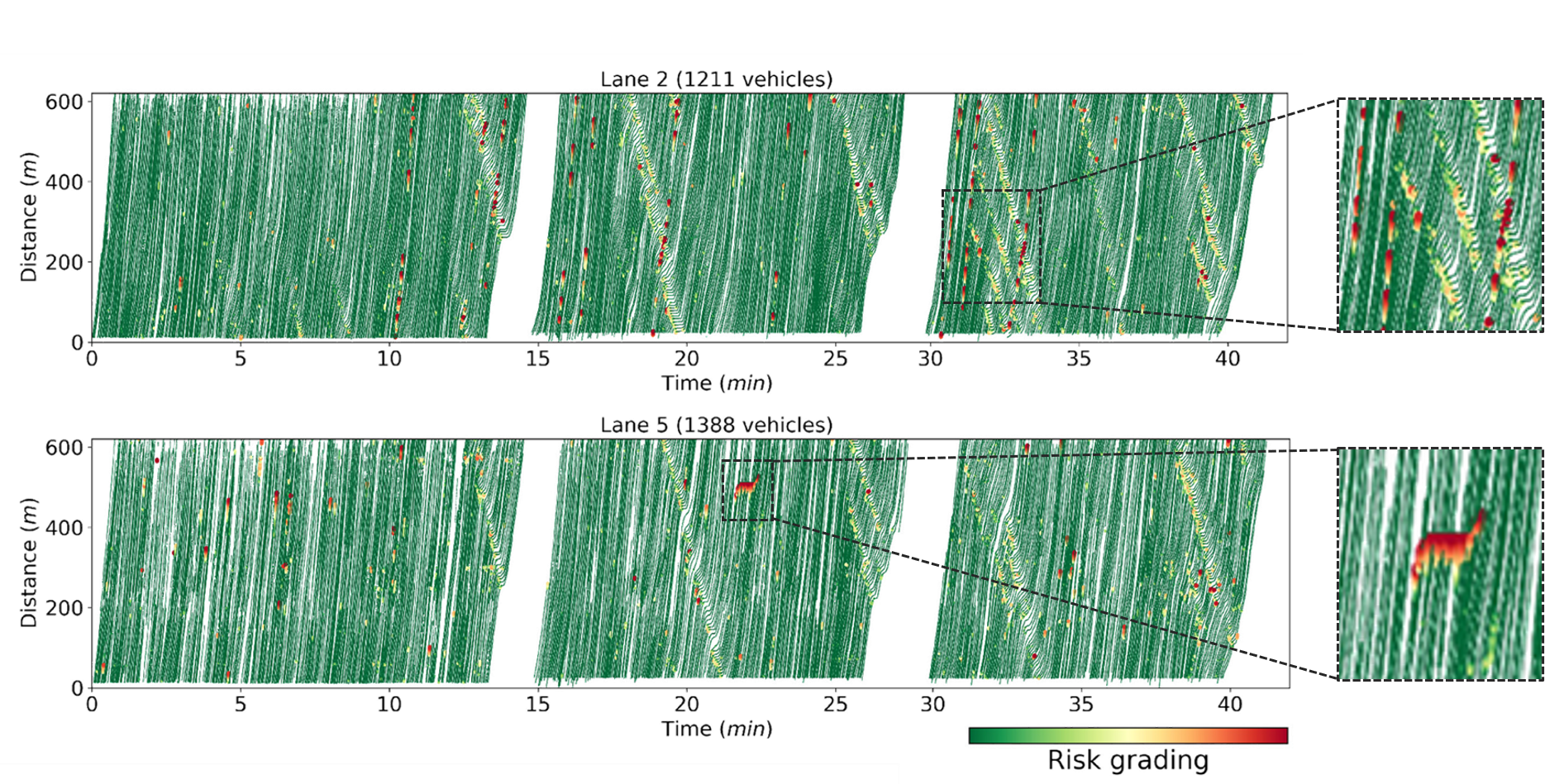}
\caption{High-resolution dynamic risk detection and profiling.}
\label{fig11}
\end{figure*}

\subsection{Learning-based Risk Profiling}
\subsubsection{Hybrid Indicators for Risk Estimation with Thresholds Calibrated by Clustering}~

The risk auto-clustering generates the optimised partitions based on risk indicators, which is also an unsupervised way to calibrate the threshold values of indicators. As highlighted by dashed horizontal lines in Fig. 10, four indicator features (i.e., $TIT.t3.max$, $TET.t1.max$, $CPI.m1.max$, $CPI.m2.max$) are obviously better in differentiating certain adjacent risk levels by simplified decision rules and thresholds, e.g., less overlapping. The 4 indicators and thresholds are listed in TABLE \ref{tab:8}, in which the thresholds are calibrated in an unsupervised way.

The 4 indicators are therefore suitable as the key signals to estimate risk potentials. $CPI$-based indicators show the ability to well isolate the higher risk levels, which is in line with the physical meanings. For multiple risk assessment, hybrid use of indicators with simplified thresholds is useful in many application scenarios (e.g., limited computation capacity, insufficient data), which is also better than relying on one stand-alone indicator with over-complicated threshold values. Besides, risk levels defined by hybrid indicators with straightforward thresholds are interpretable and easy to adopt by end-users.

\subsubsection{High-resolution Dynamic Risk Profiling}~
 
The driving risks and anomaly behaviour of a vehicle can be inferred based on the risk patterns of the belonged cluster. A high-resolution risk profiling is demonstrated in Fig. 11, in which two lanes are selected to show the microscopic risk exposures inherent in generalised traffic flow, such as the at-risk vehicles, risk severity, distribution, trends. In the plots, the $x-axis$ denotes the timestamps, in sub-second interval ($0.1s$), and the $y-axis$ denotes the travelled distance or locations on each lane, in metre-scale. 

The vehicle temporal-spatial data is integrated with dynamic risk levels, hence, target vehicles with risk potentials can be figured out, as well as the locations and timestamps with higher risk potentials. The granular risk profiling is more meaningful and useful to early identify potential crashes, and take targeted countermeasures for mitigation.

\section{Discussion}
\subsection{Limitations and Future Work}
Risk diagnosis by AutoCluster enables a predictive and proactive manner towards smart safety, which is prospected to identify and mark early risk signals and at-risk vehicles. However, result verification is a major challenge. A potential way of validation is to examine the actual crash occurrence, or the insurance claim records of the drivers/vehicles clustered to be of higher risk, which requires another way of data acquisition. 

To further improve modelling, in-depth extraction of indicator features and using high-quality data are suggested, which may involve a broader range of risk potentials and driving scenarios, such as lane changing. Methodologically, new algorithms on clustering and AutoML can be integrated to refine the solution.

In view of a large number of real-world applications suffering from the challenges of class imbalance and lack of ground truth (e.g., expensive or difficult to obtain beforehand), we demonstrate a reliable solution with a case on road safety, which holds great potentials.

\subsection{Application Potentials}
In terms of potential applications in practice, there are two aspects, one is using machine learning for data mining and knowledge discovery, and to use the found knowledge for downstream applications, another is to directly deploy the final trained model for real-time application. 

The data mining about risk exposures is clearly an area with huge value adds and also a core concern of CAV and smart road. Such application potentials are multiple folds. For roads and authorities, protective pre-crash strategies can be applied preemptively, such as patrol dispatch, flow control, road enhancement (e.g. remediation of crash-prone layouts and locations), which extends the scope of traditional policy-making and management (e.g., passive and post-accident). For vehicles and drivers, advance driver-assistance services can be provided, such as targeted risk warnings and driving recommendations (e.g. re-routing or lane-changing to avoid road segments or lanes with risk potentials).

Specifically, a real-time risk grading and diagnosis system framework is designed, as illustrated in Fig. 12. The system includes mainly three parts, offline machine learning (i.e., model training by AutoCluster), real-time prediction (i.e., model inference), as well as storage for online, nearline, and offline data and features. The offline model training task and online prediction task are asynchronous. AutoCluster learns the best-fitted models based on given data, and the final models are deployed for applications and triggered in real-time.

\begin{figure*}[!t]
\centering
\includegraphics[width=6in]{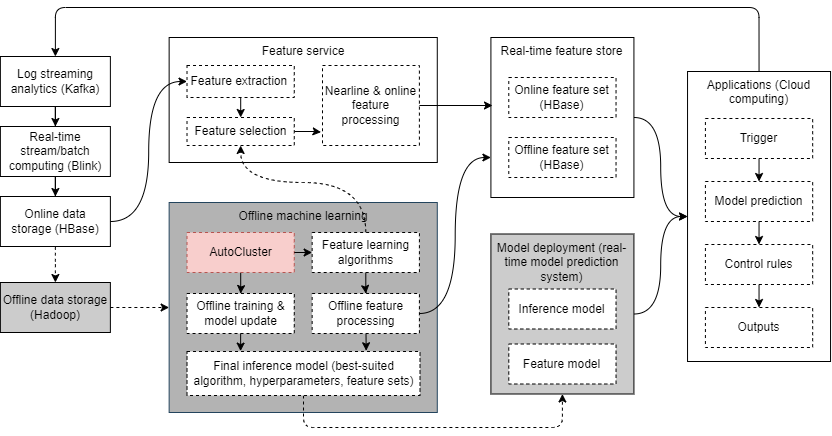}
\caption{Real-time risk diagnosis system framework.}
\label{fig12}
\end{figure*}

\section{Conclusions}

The indicator-guided automatic clustering is reliable and convincing, to end-to-end self-learn the optimal models for unsupervised risk assessment and driving anomaly detection. The developed methods are useful towards a range of advanced solutions towards smart road and crash prevention, and to tackle intrinsic challenges of clustering on highly imbalanced data without ground truth labels or $a~priori$ knowledge. In summary, this research has contributed to the domain knowledge in four areas, as highlighted in the following.

(1) Multiple risk levels in generalised driving. The AutoCluster aims to diagnose multiple distinct risk exposures inherent to generalised driving behaviour as exhibited by the vehicle stream trajectory, which is also a method of unsupervised data labelling. It is an attempt to uncover the risk exposure levels in normal traffic conditions, which extends the scope of risk analysis as relying on accidents. The method is tested on NGSIM data, and a clustering with 6 distinct risk levels is generated, as well as the risk labels per vehicle.

(2) The AutoCluster system. AutoCluster enables the auto-optimisation of risk clustering adapted onto data. Various notions of clustering algorithms and similarity measures are tested. A loss function is designed that considers the clustering performance in terms of internal quality, inter-cluster variation, and model stability. Based on Bayesian optimisation, the algorithm selection and hyperparameter tuning are self-learned, and the best risk clustering partitions with the minimum loss are generated. The performance and heterogeneity of various algorithms are also analysed.

(3) Strategies to improve imbalanced clustering. Firstly, an internal quality metric for imbalanced clustering is proposed, namely $bSI$, which is demonstrated to be more appropriate than existing CVIs. Secondly, an unsupervised feature selection method, namely EMRI, is proposed to identify the most useful features for the clustering problem. Besides, feature rectifier function is designed as a cost-sensitive strategy to improve imbalanced clustering.

(4) Learning-based risk profiling. The surrogate indicators on temporal-spatial and kinematical risk exposures are evaluated by unsupervised feature selection and ranking, and $TIT$, $TET$ and $CPI$ are found to be more useful for risk assessment. Multiple thresholds of risk indicators are bench-marked by AutoCluster. In addition, high-resolution risk detection and positioning are delineated to figure out the risk potentials in terms of targeted vehicles, locations in metre-scale, and timestamps in the sub-second interval, which enable a range of application potentials for the smart road.

% \vfill

\ifCLASSOPTIONcaptionsoff
  \newpage
\fi

\bibliographystyle{IEEEtran}
\bibliography{IEEEabrv,AC}

% Generated by IEEEtran.bst, version: 1.14 (2015/08/26)
\begin{thebibliography}{10}
\providecommand{\url}[1]{#1}
\csname url@samestyle\endcsname
\providecommand{\newblock}{\relax}
\providecommand{\bibinfo}[2]{#2}
\providecommand{\BIBentrySTDinterwordspacing}{\spaceskip=0pt\relax}
\providecommand{\BIBentryALTinterwordstretchfactor}{4}
\providecommand{\BIBentryALTinterwordspacing}{\spaceskip=\fontdimen2\font plus
\BIBentryALTinterwordstretchfactor\fontdimen3\font minus \fontdimen4\font\relax}
\providecommand{\BIBforeignlanguage}[2]{{%
\expandafter\ifx\csname l@#1\endcsname\relax
\typeout{** WARNING: IEEEtran.bst: No hyphenation pattern has been}%
\typeout{** loaded for the language `#1'. Using the pattern for}%
\typeout{** the default language instead.}%
\else
\language=\csname l@#1\endcsname
\fi
#2}}
\providecommand{\BIBdecl}{\relax}
\BIBdecl

\bibitem{eskandarian2019research}
A.~Eskandarian, C.~Wu, and C.~Sun, ``Research advances and challenges of autonomous and connected ground vehicles,'' \emph{IEEE Transactions on Intelligent Transportation Systems}, 2019.

\bibitem{Mozaffari2020}
S.~Mozaffari, O.~Y. Al-Jarrah, M.~Dianati, P.~Jennings, and A.~Mouzakitis, ``Deep learning-based vehicle behavior prediction for autonomous driving applications: A review,'' \emph{IEEE Transactions on Intelligent Transportation Systems}, pp. 1--15, 2020.

\bibitem{Zhu2018}
L.~Zhu, F.~R. Yu, Y.~Wang, B.~Ning, and T.~Tang, ``Big data analytics in intelligent transportation systems: A survey,'' \emph{IEEE Transactions on Intelligent Transportation Systems}, vol.~20, no.~1, pp. 383--398, 2018.

\bibitem{Nallaperuma2019}
D.~Nallaperuma, R.~Nawaratne, T.~Bandaragoda, A.~Adikari, S.~Nguyen, T.~Kempitiya, D.~{De Silva}, D.~Alahakoon, and D.~Pothuhera, ``Online incremental machine learning platform for big data-driven smart traffic management,'' \emph{IEEE Transactions on Intelligent Transportation Systems}, vol.~20, no.~12, pp. 4679--4690, 2019.

\bibitem{shi2019feature}
X.~Shi, Y.~D. Wong, M.~Z.-F. Li, C.~Palanisamy, and C.~Chai, ``A feature learning approach based on xgboost for driving assessment and risk prediction,'' \emph{Accident Analysis and Prevention}, vol. 129, pp. 170--179, 2019.

\bibitem{Shi2020}
X.~Shi, Y.~D. Wong, C.~Chai, and M.~Z.-F. Li, ``An automated machine learning (automl) method of risk prediction for decision-making of autonomous vehicles,'' \emph{IEEE Transactions on Intelligent Transportation Systems}, 2020.

\bibitem{Shi2018}
X.~Shi, Y.~D. Wong, M.~Z.~F. Li, and C.~Chai, ``Key risk indicators for accident assessment conditioned on pre-crash vehicle trajectory,'' \emph{Accident Analysis and Prevention}, vol. 117, pp. 346--356, 2018.

\bibitem{Chai2015}
C.~Chai and Y.~D. Wong, ``Fuzzy cellular automata model for signalized intersections,'' \emph{Computer-Aided Civil and Infrastructure Engineering}, vol.~30, no.~12, pp. 951--964, 2015.

\bibitem{chai2014micro}
C.~Chai and Y.~Wong, ``Micro-simulation of vehicle conflicts involving right-turn vehicles at signalized intersections based on cellular automata,'' \emph{Accident Analysis and Prevention}, vol.~63, pp. 94--103, 2014.

\bibitem{Perez}
M.~A. Perez, J.~D. Sudweeks, E.~Sears, J.~Antin, S.~Lee, J.~M. Hankey, and T.~A. Dingus, ``Performance of basic kinematic thresholds in the identification of crash and near-crash events within naturalistic driving data,'' \emph{Accident Analysis and Prevention}, vol. 103, pp. 10--19, 2017.

\bibitem{Siami2020}
M.~Siami, M.~Naderpour, and J.~Lu, ``A mobile telematics pattern recognition framework for driving behavior extraction,'' \emph{IEEE Transactions on Intelligent Transportation Systems}, pp. 1--14, 2020.

\bibitem{Zheng2014}
L.~Zheng, K.~Ismail, and X.~Meng, ``Traffic conflict techniques for road safety analysis: Open questions and some insights,'' \emph{Canadian Journal of Civil Engineering}, vol.~41, no.~7, pp. 633--641, 2014.

\bibitem{Mahmud2017}
S.~S. Mahmud, L.~Ferreira, M.~S. Hoque, and A.~Tavassoli, ``Application of proximal surrogate indicators for safety evaluation: A review of recent developments and research needs,'' \emph{IATSS Research}, vol.~41, no.~4, pp. 153--163, 2017.

\bibitem{Laureshyn}
A.~Laureshyn, {\AA}.~Svensson, and C.~Hyd{\'e}n, ``Evaluation of traffic safety, based on micro-level behavioural data: Theoretical framework and first implementation,'' \emph{Accident Analysis and Prevention}, vol.~42, no.~6, pp. 1637--1646, 2010.

\bibitem{Estivill-Castro2002}
V.~Estivill-Castro, ``Why so many clustering algorithms,'' \emph{ACM SIGKDD Explorations Newsletter}, vol.~4, no.~1, pp. 65--75, 2002.

\bibitem{Rodriguez2014}
A.~Rodriguez and A.~Laio, ``Clustering by fast search and find of density peaks,'' \emph{Science}, vol. 344, no. 6191, pp. 1492--1496, 2014.

\bibitem{elkan2001foundations}
C.~Elkan, ``The foundations of cost-sensitive learning,'' in \emph{International Joint Conference on Artificial Intelligence}, vol.~17, no.~1, 2001, pp. 973--978.

\bibitem{Diez-Pastor2015}
J.~F. D{\'{i}}ez-Pastor, J.~J. Rodr{\'{i}}guez, I.~Garc{\'{i}}a-Osorio, and I.~Kuncheva, ``Diversity techniques improve the performance of the best imbalance learning ensembles,'' \emph{Information Sciences}, vol. 325, pp. 98--117, 2015.

\bibitem{Beyan}
C.~Beyan and R.~Fisher, ``Classifying imbalanced data sets using similarity based hierarchical decomposition,'' \emph{Pattern Recognition}, vol.~48, no.~5, pp. 1653--1672, 2015.

\bibitem{Lopez2013}
V.~L{\'{o}}pez, A.~Fern{\'{a}}ndez, S.~Garc{\'{i}}a, V.~Palade, and F.~Herrera, ``An insight into classification with imbalanced data: Empirical results and current trends on using data intrinsic characteristics,'' \emph{Information Sciences}, vol. 250, pp. 113--141, 2013.

\bibitem{de2015recovering}
R.~C. de~Amorim and C.~Hennig, ``Recovering the number of clusters in data sets with noise features using feature rescaling factors,'' \emph{Information Sciences}, vol. 324, pp. 126--145, 2015.

\bibitem{Van}
T.~van Craenendonck and H.~Blockeel, ``Using internal validity measures to compare clustering algorithms,'' \emph{Journal of Computational and Applied Mathematics}, pp. 1--8, 2015.

\bibitem{shahriari2015taking}
B.~Shahriari, K.~Swersky, Z.~Wang, R.~P. Adams, and N.~De~Freitas, ``Taking the human out of the loop: A review of bayesian optimization,'' \emph{Proceedings of the IEEE}, vol. 104, no.~1, pp. 148--175, 2015.

\bibitem{Snoek}
J.~Snoek, H.~Larochelle, and R.~P. Adams, ``Practical bayesian optimization of machine learning algorithms,'' in \emph{Advances in Neural Information Processing Systems}, 2012, pp. 2951--2959.

\bibitem{Bergstra}
J.~S. Bergstra, R.~Bardenet, Y.~Bengio, and B.~K{\'e}gl, ``Algorithms for hyper-parameter optimization,'' in \emph{Advances in Neural Information Processing Systems}, 2011, pp. 2546--2554.

\bibitem{Rousseeuw1987}
P.~J. Rousseeuw, ``Silhouettes: a graphical aid to the interpretation and validation of cluster analysis,'' \emph{Journal of Computational and Applied Mathematics}, vol.~20, pp. 53--65, 1987.

\bibitem{Bousquet2002}
O.~Bousquet and A.~Elisseeff, ``Stability and generalization,'' \emph{Journal of Machine Learning Research}, vol.~2, pp. 499--526, 2002.

\bibitem{Guyon2003}
I.~Guyon and A.~Elisseeff, ``An introduction to variable and feature selection,'' \emph{Journal of Machine Learning Research}, vol.~3, pp. 1157--1182, 2003.

\bibitem{Horst1991}
A.~van~der Horst, ``A time-based analysis of road user behaviour in normal and critical encounters,'' Ph.D. dissertation, Delft University of Technology, 1990.

\bibitem{Minderhoud}
M.~M. Minderhoud and P.~H. Bovy, ``Extended time-to-collision measures for road traffic safety assessment,'' \emph{Accident Analysis and Prevention}, vol.~33, no.~1, pp. 89--97, 2001.

\bibitem{Archer2005}
J.~Archer, ``Indicators for traffic safety assessment and prediction and their application in micro-simulation modelling: A study of urban and suburban intersections,'' Ph.D. dissertation, KTH, 2005.

\bibitem{Cunto}
F.~Cunto and F.~F. Saccomanno, ``Calibration and validation of simulated vehicle safety performance at signalized intersections,'' \emph{Accident Analysis and Prevention}, vol.~40, no.~3, pp. 1171--1179, 2008.

\bibitem{Zarshenas}
A.~Zarshenas and K.~Suzuki, ``Binary coordinate ascent: An efficient optimization technique for feature subset selection for machine learning,'' \emph{Knowledge-Based Systems}, vol. 110, pp. 191--201, 2016.

\bibitem{Fisher2019}
A.~Fisher, C.~Rudin, and F.~Dominici, ``All models are wrong, but many are useful: Learning a variable's importance by studying an entire class of prediction models simultaneously,'' \emph{Journal of Machine Learning Research}, vol.~20, no. 177, pp. 1--81, 2019.

\bibitem{halkias2006next}
J.~Halkias and J.~Colyar, ``Next generation simulation fact sheet,'' \emph{US Department of Transportation: Federal Highway Administration}, 2006.

\bibitem{Punzo}
V.~Punzo, M.~T. Borzacchiello, and B.~Ciuffo, ``On the assessment of vehicle trajectory data accuracy and application to the next generation simulation (ngsim) program data,'' \emph{Transportation Research Part C: Emerging Technologies}, vol.~19, no.~6, pp. 1243--1262, 2011.

\end{thebibliography}

\begin{IEEEbiography}[{\includegraphics[width=1in,height=1.25in,clip,keepaspectratio]{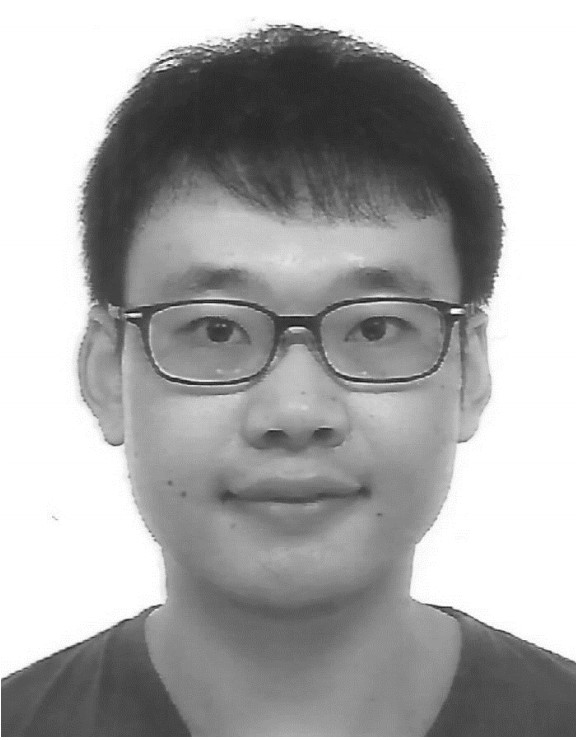}}]{Xiupeng Shi}
received the B.Eng. and M.Eng. degrees from Beijing Jiaotong University, China, in 2010 and 2012, respectively, and the Ph.D. degree from Nanyang Technological University (NTU), Singapore, in 2019. 

He is currently an Algorithm Expert with Alipay, Alibaba and Ant Group. From 2020 to 2021, he was a Scientist with Agency for Science, Technology and Research Singapore (A*STAR). From 2012 to 2015, he worked as a Business General Manager with China Railway Materials. His research interests include risk prediction, automated machine learning, recommender system, smart mobility, and supply chain finance. 
\end{IEEEbiography}

\begin{IEEEbiography}[{\includegraphics[width=1in,height=1.25in,clip,keepaspectratio]{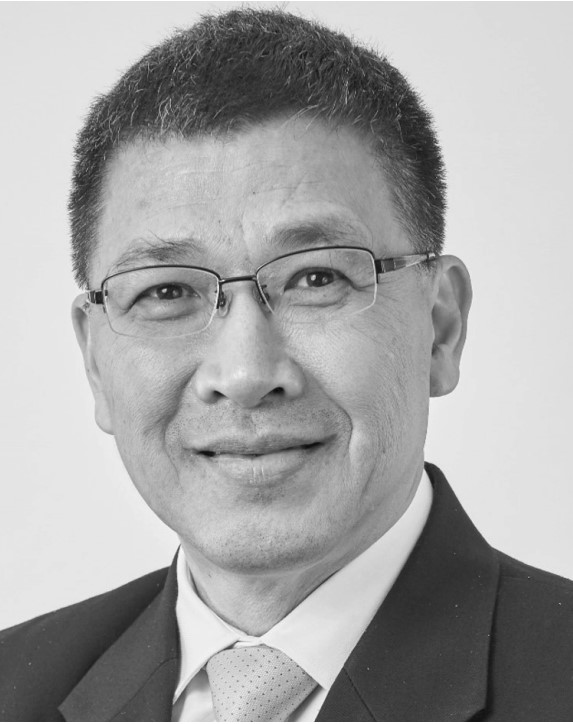}}]{Yiik Diew Wong}
received the B.E. degree in civil engineering and the Ph.D. degree in transportation engineering from the University of Canterbury, New Zealand, in 1983 and 1990, respectively.

He is currently an Associate Chair (Academic) and an Associate Professor with the School of Civil and Environmental Engineering, Nanyang Technological University (NTU), Singapore. His research interests include sustainable urban mobility, road safety engineering, driver behaviors, and active mobility.
\end{IEEEbiography}

% \vfill

\begin{IEEEbiography}[{\includegraphics[width=1in,height=1.25in,clip,keepaspectratio]{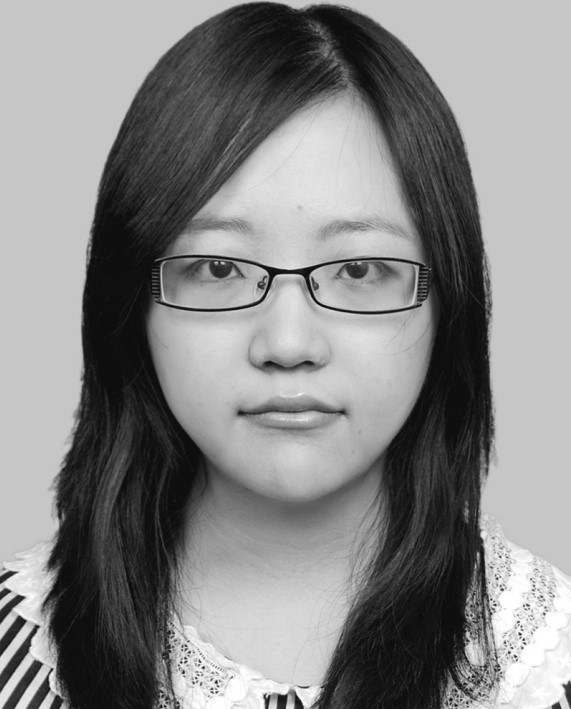}}]{Chen Chai}
(Member, IEEE) received the B.Eng. degree from Beijing Jiaotong University, China, in 2010, the M.Sc. degree from The Hong Kong University of Science and Technology, Hong Kong, in 2011, and the Ph.D. degree from Nanyang Technological University, Singapore, in 2015. 

She is currently an Associate Professor with the College of Transportation Engineering, Tongji University, China. Her research interests include human-vehicle cooperative driving systems and mixed traffic flow of manual vehicles and autonomous vehicles. She is also a member of the Transportation Research Board Standing Committee ACH50: Human Factors and Behavioral Research Methods. She was a recipient of the Shanghai Sailing Talent Award and the Shanghai Chenguang Talent Award in 2018. 
\end{IEEEbiography}

\begin{IEEEbiography}[{\includegraphics[width=1in,height=1.25in,clip,keepaspectratio]{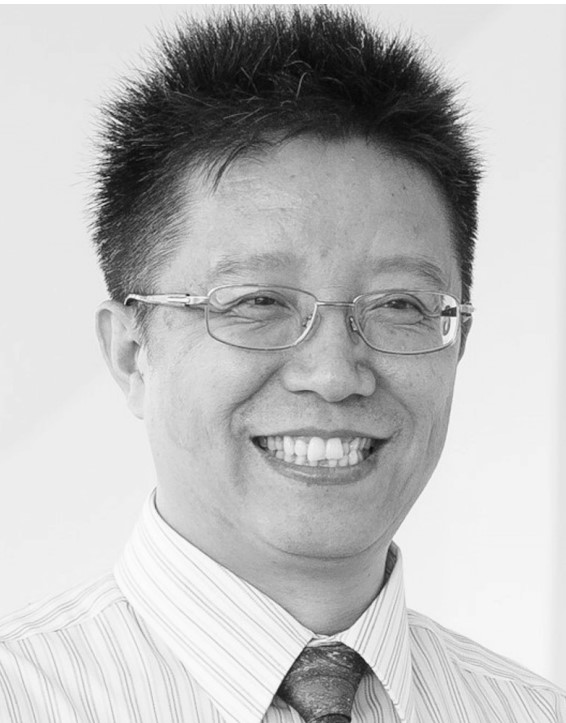}}]{Michael Zhi-Feng Li}
received the B.Sc. degree in mathematics from Beijing Normal University, China, in 1983, the Ph.D. degree in mathematics from the University of Regina, Canada, in 1989, and the Ph.D. degree in business from The University of British Columbia, Canada, in 1994. 

He is currently an Associate Professor with the Nanyang Business School, Nanyang Technological University (NTU), Singapore. He is also the Director of the Future China Advanced Leadership Programme and the Nanyang Executive MBA Programme in NTU. His research interests include operations research, transport economics, revenue management, and network congestion pricing. 
\end{IEEEbiography}

\begin{IEEEbiography}[{\includegraphics[width=1in,height=1.25in,clip,keepaspectratio]{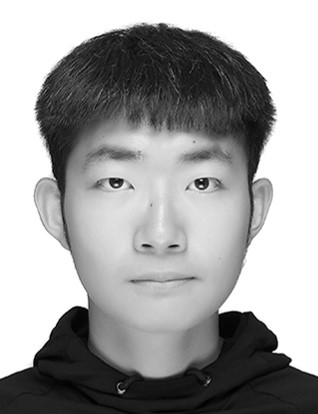}}]{Tianyi Chen}
received the B.Eng. degree in logistics engineering from Wuhan University of Technology, China, in 2017. He graduated his the M.Sc. degree in supply chain and logistics management and the Ph.D. degree in transportation engineering and data science from Nanyang Technological University (NTU), Singapore, in 2018 and 2021, respectively.

He is currently a research fellow at the School
of Civil and Environmental Engineering, NTU. His
research interests include traffic safety analysis, logistics management, data science, and statistical learning.

\end{IEEEbiography}

\begin{IEEEbiography}[{\includegraphics[width=1in,height=1.25in,clip,keepaspectratio]{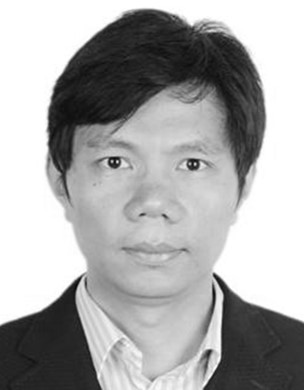}}]{Zeng Zeng}
(Senior Member, IEEE) received the Ph.D. degree in electrical and computer engineering from the National University of Singapore, Singapore, in 2004. 

He currently works as a Senior Scientist and Program Head with I2R, A*Star, Singapore. From 2011 to 2014, he was a Senior Research Fellow with the National University of Singapore. From 2005 to 2011, he was a Professor with the Computer and Communication School, Hunan University, Changsha, China. His research interests include distributed/ parallel computing systems, data stream analysis, deep learning, multimedia storage systems, and wireless sensor networks.
\end{IEEEbiography}

\vfill

\end{document}